\documentclass[10pt,twocolumn]{article}

\newif\ifarxiv
\arxivtrue

\usepackage[utf8]{inputenc}
\usepackage[T1]{fontenc}
\usepackage[letterpaper,top=0.66in,bottom=0.66in,left=0.65in,right=0.65in,columnsep=0.24in]{geometry}
\usepackage{times}
\usepackage{microtype}
\usepackage{amsmath,amssymb,amsthm}
\usepackage{graphicx}
\usepackage{booktabs}
\usepackage{multirow}
\usepackage{xcolor}
\usepackage{enumitem}
\usepackage{algorithm}
\usepackage{algpseudocode}
\usepackage[numbers,sort&compress]{natbib}
\usepackage[hidelinks]{hyperref}
\usepackage{caption}
\usepackage{placeins}
\usepackage{dblfloatfix}
\usepackage{adjustbox}

\setlist{nosep,leftmargin=1.3em}
\newtheorem{proposition}{Proposition}
\theoremstyle{definition}
\newtheorem{definition}{Definition}

\newcommand{\sys}{\textsc{CommitJIT}}

\ifarxiv
\hypersetup{
  pdftitle={Screenshots Are Not State: Transactional Verification and Compilation for GUI Agents},
  pdfauthor={Richard Abrich}
}
\fi

\begin{document}

\twocolumn[{
\begin{center}
\rule{\linewidth}{2.0pt}\\[0.5em]
{\Large\bfseries Screenshots Are Not State:\\[0.12em] Transactional Verification and Compilation for GUI Agents}\\[0.4em]
\rule{\linewidth}{0.7pt}\\[0.65em]
\ifarxiv
{\normalsize\bfseries Richard Abrich}\\[0.2em]
{\small OpenAdapt.AI (MLDSAI Inc.), Toronto, Canada}\\
{\small \texttt{richard@openadapt.ai} \quad$\cdot$\quad ORCID: 0000-0002-9556-4491}\\[0.3em]
{\footnotesize Preprint v2 (December 2026). Code: \url{https://github.com/OpenAdaptAI/OpenAdapt} and \url{https://github.com/OpenAdaptAI/openadapt-evals}.}
\else
{\normalsize\bfseries Anonymous Authors}\\
{\small Anonymous Institution}\\
{\small \texttt{anonymous@example.com}}\\[0.3em]
{\footnotesize Double-blind submission. Under review.}
\fi
\end{center}
\vspace{0.1em}
}]

\begin{abstract}
GUI agents still fail silently on multi-step enterprise tasks, reporting completion when a form was never submitted, a required tab was missed, or a policy was violated. A vision-language critic before every step backfires: per-step false positives compound with horizon length, and one screenshot cannot see off-screen tabs, closed folders, or writes that no view shows. We introduce \sys{}, a runtime wrapper that checks only the few state-changing and policy boundaries ($K \ll T$). It combines deterministic before-commit checks, a write-ahead saga log, graded post-write read-back, a calibrated local 4B visual verifier, and verified replay that repairs across UI upgrades with zero raw field-value egress. \sys{} cuts Silent Error Rate (SER) without degrading completion on 3-app \texttt{ST-WebAgentBench} (95.5\% $\rightarrow$ 27.3\%, $N=44$, $p=1.9\times 10^{-9}$, CuP 4.5\% $\rightarrow$ 6.8\%), \texttt{OS-Harm} (35.0\% $\rightarrow$ 5.0\%, $N=20$, $p=0.031$, Verified Completion Rate (VCR) 60.0\% both), \texttt{ERPBench} (31.3\% $\rightarrow$ 0.0\%, $N=16$, $p=0.0625$, VCR 68.8\% $\rightarrow$ 81.3\%), and 60 real Docker upgrades (93.3\% autonomous and 100.0\% zero-egress survival at 0.0\% SER). Where default-deny trades coverage for safety, \texttt{HealthAdminBench} SER falls from 61.5\% to 1.9\% ($N=52$, $p=9.3\times 10^{-10}$; VCR 38.5\% $\rightarrow$ 11.5\%) and from 32.6\% to 17.8\% on the full 135-task zero-SOP suite ($p=5.4\times 10^{-4}$; VCR 6.7\% $\rightarrow$ 3.7\%). On a 32-episode worker-crash rerun where every arm sees the page text, all arms finish exactly once with zero duplicates; the wrapper's gain is 100 vs.\ 280 model calls. Boundaries remain for unwritten clinic rules, clinical judgment (\texttt{MedHarnessBench} $N=12$: 0.0\% SER at 0.0\% VCR, abstaining 12/12), non-visual desktop state, and compressed remote-desktop streams. This version corrects a v1 crash-duplicate result whose baseline could not see page text.
\end{abstract}

\section{Introduction}
\label{sec:intro}

\begin{figure*}[t]
\centering
\includegraphics[width=0.93\textwidth]{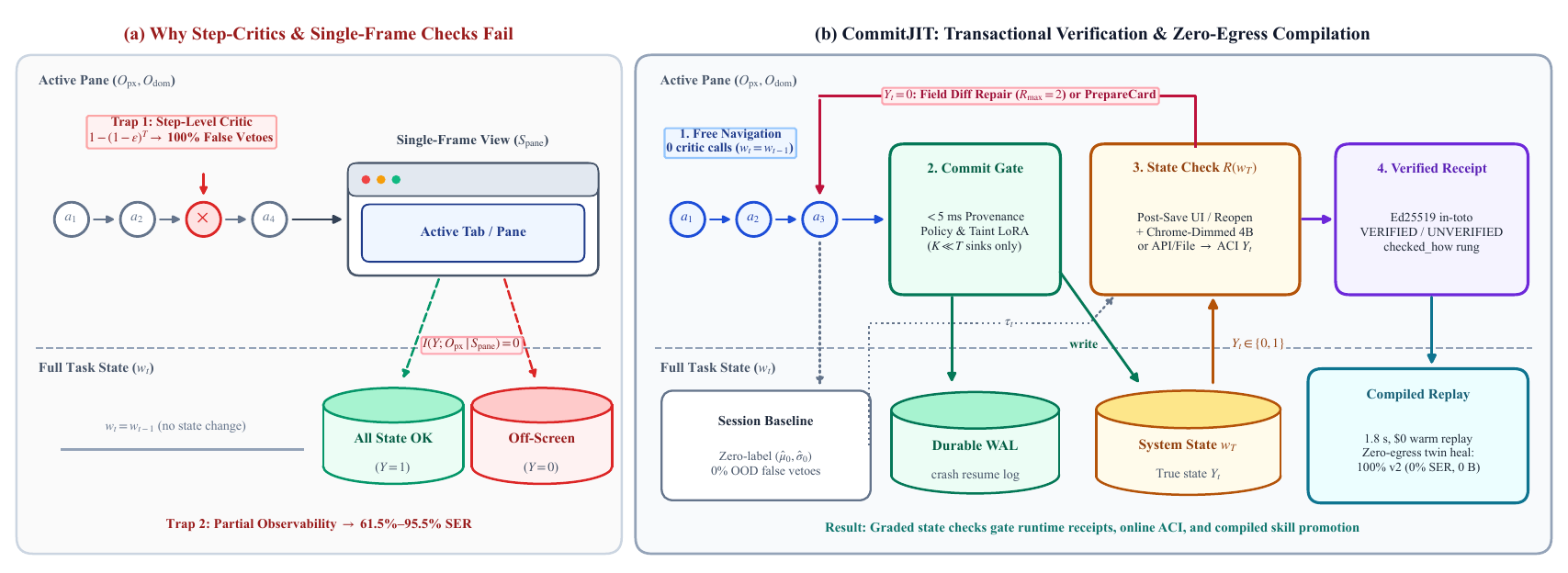}
\caption{\textbf{Screenshots are not state: problem diagnosis and \sys{} architecture.} \textbf{(a)} Step-level critics ($w_t = w_{t-1}$) compound false vetoes as $1-(1-\epsilon)^T$, while a single active-pane screenshot ($O_{\text{px}}$) cannot observe off-screen tabs, closed folders, config files, or unlisted writes in full task state $w_T$ ($I(Y;O_{\text{px}}\mid S_{\text{pane}})=0$). \textbf{(b)} \sys{} leaves navigation unchecked to prime $(\hat{\mu}_0,\hat{\sigma}_0)$, gates only the $K \ll T$ commits with a $<5$\,ms check and durable WAL, and verifies state via a graded ladder $R(w_T)$ to supervise online ACI, sign receipts, and gate zero-egress compiled replays.}
\label{fig:arch}
\end{figure*}

Frontier vision-language models (VLMs) and specialized computer-use models~\cite{xie2024osworld,zhou2024webarena,qin2025uitars,wu2025osatlas,xu2024aguvis} now navigate complex web and desktop interfaces, yet they still fail quietly in healthcare, insurance, and enterprise back-office workflows. Recent diagnostic benchmarks document why: long-horizon control mistakes silently corrupt downstream steps (\textsc{CUADebug}~\cite{cuadebug2026}), dense controls cause mis-clicks (\textsc{ComponentBench}~\cite{componentbench2026}, \textsc{Trustworthy GUI Agents}~\cite{trustworthygui2025}), multi-tab EHR and payer portals drift out of sync (\textsc{HealthAdminBench}~\cite{healthadminbench2026}, \textsc{MedHarnessBench}~\cite{medharnessbench2026}), and agents report completion while skipping user consent (\textsc{ST-WebAgentBench}~\cite{levy2024stwebagentbench}) or overriding harms they explicitly noticed in their own reasoning (\textsc{SEAM}~\cite{seam2026}).

In regulated settings, error costs are asymmetric. Pausing visibly to ask an operator for sign-off costs seconds; a \emph{silent error}, where an agent claims success while the backing system holds a missing, wrong, duplicate, or unauthorized record, creates clinical and financial liability.

A common response is to place a VLM critic in front of every action. We show mathematically and empirically why per-step visual gating fails on long-horizon workflows:
\begin{enumerate}
  \item \textbf{False-veto compounding (Proposition~\ref{prop:compounding}).} Over a $T$-step episode, a per-step critic with false-positive rate $\epsilon > 0$ triggers at least one false abort with probability $1 - (1-\epsilon)^T$. At $\epsilon = 0.05$ and $T = 25$, false aborts reach 72.3\%; under out-of-distribution UI shift, completion drops to zero. Restricting checks to the $K \ll T$ irreversible commit sinks bounds cumulative false vetoes by $K\alpha$ without losing recall on backend state errors.
  \item \textbf{Single-frame partial observability (Proposition~\ref{prop:unidentifiable}).} A single screenshot of the active pane carries zero mutual information with state that lives off-screen (such as an unvisited EHR tab, a closed mail folder, a background configuration file, or a saved record that no list or search view shows). Neither a human nor a pixel-only classifier looking only at that frame can verify completion without inspecting the relevant view, reading back the target state, or checking a durable log.
\end{enumerate}

\paragraph{Overview of \sys{}.}
We build \sys{} (Figure~\ref{fig:arch}), a model-agnostic runtime wrapper that concentrates verification at policy and commit boundaries ($K \ll T$). First, a deterministic before-commit layer ($<5$\,ms) checks entity provenance, policy tables, accessibility grammar, local prompt-injection taint, and a disk-backed write-ahead saga log (WAL) that prevents duplicate writes across crashes and re-authentication drops. Second, at commit sinks, \sys{} uses a graded post-write verification ladder (starting with a post-save screen check as the universal baseline, escalating to reopened views, separate read-only sessions, or optional API/database checks when the target lives off-screen or the app supports them), backed by a chrome-normalized 4B local visual verifier when DOM structure is absent or ambiguous. Third, \sys{} separates zero-label session-start normalization on benign navigation frames from online Adaptive Conformal Inference (ACI)~\cite{gibbs2021aci}, which updates only when verification resolves a ground-truth label. Fourth, verified traces compile into deterministic replays that re-ground against live DOM indexes on UI upgrades and repair structural breaks inside a bijectively symbolized offline twin with zero raw field-value egress.

\paragraph{Contributions.}
\begin{enumerate}
  \item \textbf{Formal diagnosis:} We formalize false-veto compounding and single-frame partial observability (Section~\ref{sec:theory}) and validate both directly on real desktop trajectories.
  \item \textbf{Commit-boundary runtime:} We introduce \sys{}, integrating a three-level graded verification ladder, a crash-safe write-ahead saga log, regime-matched calibration, and zero-egress self-healing compilation (Section~\ref{sec:method}).
  \item \textbf{External multi-suite evaluation:} We test \sys{} on external web, clinical, ERP, IT-service, safety, and desktop benchmarks, real upstream Docker software upgrades, and real Windows H.264 remote-desktop runs, reporting VCR beside SER with exact 95\% Wilson confidence intervals, exact two-sided McNemar tests, and a mechanism-to-suite summary mapping each component to its isolating benchmark (Section~\ref{sec:experiments}).
  \item \textbf{Empirical boundary map:} Rather than claiming a universal oracle, we document where the wrapper stops helping: unwritten institutional policies without demonstrations, non-visual desktop state errors without read-back, conservative escalation tradeoffs, pixel jitter on compressed remote-desktop streams, and single-worker crash recovery, where a frontier model that can read the page text already avoids duplicates (Sections~\ref{sec:experiments} and~\ref{sec:limits}).
\end{enumerate}

\ifarxiv
\paragraph{Changes in v2.}
Version~1 reported that after a mid-batch worker crash, unwrapped agents duplicated a record in 10 of 16 (\texttt{gpt-6.2-sol}) and 11 of 16 (\texttt{gpt-5.6-sol}) episodes against 0 of 16 wrapped. That baseline gave the agent only form controls and links, with no page text and no memory of what it had seen, so it often could not see the record it had already saved. A rerun that gives every arm the same visible page text and memory finds 0 of 32 duplicates in all arms (Section~\ref{sec:rq4}). We retract the v1 duplicate gap, drop the pooled 64-episode duplicate claim that included it, and add six paired live runs on external suites, one of which replaces an earlier \texttt{OS-Harm} wrapper result on the same 20 tasks (Section~\ref{sec:wave2}).
\fi

\section{Problem Formulation and Two Structural Barriers}
\label{sec:theory}

Consider a GUI agent executing a trajectory $\tau = (s_0, a_1, s_1, \dots, a_T, s_T)$ of horizon $T$ with latent backend state $w_t \in \mathcal{W}$ and observation $o_t = (o_t^{\text{px}}, o_t^{\text{dom}}) \in \mathcal{O}$. Actions partition into reversible read-only navigation $\mathcal{A}_{\text{nav}}$ ($w_t = w_{t-1}$) and state-mutating commit sinks $\mathcal{A}_{\text{commit}}$ (submitting a form, saving a clinical record, sending a message, executing a write). Let $\mathcal{I}_{\text{commit}} = \{t \in [T] : a_t \in \mathcal{A}_{\text{commit}}\}$ and $K = |\mathcal{I}_{\text{commit}}|$. Typical enterprise workflows span $T \in [15, 60]$ actions with $K \in [1, 3]$ commits, so $K \ll T$.

Each task specifies true backend completion $G(w_T) \in \{0,1\}$ and policy compliance $\Pi(\tau) \in \{0,1\}$. A runtime verifier outputs $V(\tau) \in \{\texttt{VERIFIED}, \texttt{VETO}, \texttt{ABSTAIN}\}$, where \texttt{VETO} requests bounded repair and \texttt{ABSTAIN} halts for human sign-off.

\begin{definition}[Verification Metrics]
\label{def:metrics}
Across $M$ tasks, let $C_i = G(w_T^{(i)})$, $P_i = 1 - \Pi(\tau_i)$, $E_i = \mathbb{I}[V(\tau_i) = \texttt{ABSTAIN}]$, and $A_i = \mathbb{I}[V(\tau_i) = \texttt{VERIFIED}]$. We track Verified Completion Rate (VCR), Silent Error Rate (SER), and Escalation Precision (HEP):
\begin{align}
  \text{VCR} &= \frac{1}{M}\sum_{i=1}^M A_i C_i (1 - P_i), \\
  \text{SER} &= \frac{1}{M}\sum_{i=1}^M A_i \bigl(1 - C_i(1 - P_i)\bigr), \\
  \text{HEP} &= \frac{\sum_{i=1}^M E_i \bigl(1 - C_i(1 - P_i)\bigr)}{\sum_{i=1}^M E_i},
\end{align}
alongside raw task completion $\frac{1}{M}\sum C_i$, policy violation rate $\frac{1}{M}\sum P_i$, and policy-compliant completion ($\text{CuP} = \frac{1}{M}\sum C_i(1-P_i)$).
\end{definition}

An unwrapped agent implicitly sets $A_i = 1$ on every finished run, converting every failed or policy-violating run into a silent error. Note that a verifier can push SER to 0\% trivially by abstaining on every task ($A_i = 0$); VCR and HEP therefore must always be read beside SER.

\begin{proposition}[False-Veto Compounding vs.\ Sink-Only Gating]
\label{prop:compounding}
Consider a $T$-step trajectory with $K \ll T$ irreversible commit sinks ($w_t = w_{t-1}$ for all $t \notin \mathcal{I}_{\text{commit}}$).
(a)~Any step-level critic with independent per-step false-positive rate at least $\epsilon > 0$ on benign steps incurs cumulative false-veto probability
\begin{equation}
  P_{\text{step}}(\text{false veto}) \ge 1 - (1 - \epsilon)^T,
\end{equation}
which approaches $1$ exponentially in $T$ (Figure~\ref{fig:compounding}(a)).
(b)~Gating strictly on the $K$ commit sinks at per-commit false-positive rate $\alpha$ bounds cumulative false vetoes by
\begin{equation}
  P_{\text{sink}}(\text{false veto}) = 1 - (1 - \alpha)^K \le K\alpha,
\end{equation}
reducing false aborts by factor $T/K$ for small $\epsilon = \alpha$ (and to $0$ for deterministic checks) while losing zero recall on irreversible backend state errors relative to gating all $T$ steps, because every change to $w_T$ passes through $\mathcal{I}_{\text{commit}}$.
\end{proposition}

\begin{proposition}[Single-Frame Partial Observability]
\label{prop:unidentifiable}
Let $Y = G(w_T) \in \{0,1\}$ indicate whether full task state $w_T$ satisfies the target postcondition, and let $O_{\text{px}}$ be a single screenshot of the active pane with visible UI state $S_{\text{pane}}$. Whenever the determining component of $w_T$ is off-screen or outside the active worker's memory (for instance, an unvisited EHR tab, a closed mail folder or configuration file on desktop, a write saved by a concurrent worker on the same job, or a record that no list or search view shows), $O_{\text{px}} \perp\!\!\!\perp Y \mid S_{\text{pane}}$ and $I(Y; O_{\text{px}} \mid S_{\text{pane}}) = 0$. By the data processing inequality, for any single-frame pixel verifier $f(O_{\text{px}})$,
\begin{equation}
\begin{aligned}
  &\Pr\bigl(f(O_{\text{px}}) = \texttt{PASS},\, Y = 0 \mid S_{\text{pane}}\bigr) \\
  &\quad = \Pr\bigl(f(O_{\text{px}}) = \texttt{PASS} \mid S_{\text{pane}}\bigr)\Pr(Y = 0 \mid S_{\text{pane}}).
\end{aligned}
\end{equation}
No single-frame pixel verifier can combine positive acceptance with zero SER unless paired with an inspection channel $R(w_T)$ (inspecting the relevant view, reopening or querying the saved record, or checking a durable WAL) satisfying $I(Y; R(w_T) \mid S_{\text{pane}}) > 0$.
\end{proposition}

Proofs appear in Appendix~\ref{app:proofs}.

\section{Method: \sys{}}
\label{sec:method}

Figure~\ref{fig:arch}(b) and Algorithm~\ref{alg:commitjit} summarize \sys{}. Read-only navigation runs without visual critics while priming a baseline distribution of benign frames. State-mutating actions pass through deterministic before-commit checks, a disk-backed write-ahead saga log, graded post-write read-back, and bounded repair.

\begin{algorithm}[t]
\caption{\sys{} Commit-Boundary Execution Loop}
\label{alg:commitjit}
\footnotesize
\begin{algorithmic}[1]
\Require Agent $\pi$, policy table $\mathcal{P}$, dossier $D$, WAL $\mathcal{L}$, max repairs $R_{\max}=2$
\State Load committed item keys $\mathcal{K}_{\text{done}} \leftarrow \text{ReplayWAL}(\mathcal{L})$ \Comment{Survives crashes}
\State Initialize session-start baseline $(\hat{\mu}_0, \hat{\sigma}_0)$ over first $K_{\text{nav}}$ benign frames
\For{$t = 1, \dots, T$}
  \State Propose action $a_t \leftarrow \pi(o_t, D, \mathcal{K}_{\text{done}})$
  \If{$a_t \in \mathcal{A}_{\text{nav}}$} \Comment{Read-only navigation}
    \State Execute $a_t$; update $(\hat{\mu}_0, \hat{\sigma}_0)$ if $t \le K_{\text{nav}}$; \textbf{continue}
  \EndIf
  \State \textbf{// Level 1: Deterministic before-commit guard ($<5$\,ms)}
  \If{Deduplicated($a_t, \mathcal{K}_{\text{done}}$) \textbf{or} Violates($a_t, \mathcal{P}, D$) \textbf{or} Tainted($a_t$)}
    \State \textbf{return} \texttt{ABSTAIN} (emit \texttt{PrepareCard} or skip committed item)
  \EndIf
  \State Append \texttt{PENDING}$(a_t)$ to durable WAL $\mathcal{L}$; execute $a_t$
  \State \textbf{// Levels 2--3: Graded post-write read-back}
  \State $(Y_t, \Delta_t, \text{rung}) \leftarrow \text{GradedReadBack}(o_{t+1}, D, \tau_t)$
  \If{$Y_t = 1$}
    \State Append \texttt{COMMITTED}$(a_t)$ to $\mathcal{L}$; update ACI $\alpha_{t+1}$ from $Y_t$
  \ElsIf{repairs used $< R_{\max}$ and $\Delta_t \neq \emptyset$}
    \State Feed structured field diff $\Delta_t$ back to $\pi$ for repair
  \Else
    \State \textbf{return} \texttt{ABSTAIN} (route staged \texttt{PrepareCard} to operator)
  \EndIf
\EndFor
\State \textbf{return} \texttt{VERIFIED} + Ed25519 receipt ($\text{checked\_how} = \min\text{rung}$)
\end{algorithmic}
\end{algorithm}

\subsection{Three-Level Graded Verification Ladder}

\paragraph{Level 1: Deterministic before-commit guard ($<5$\,ms, 0\% false vetoes).}
Before executing any state-mutating action, \sys{} runs four fast checks (module details in Appendix~\ref{app:proofs}):
\begin{enumerate}
  \item \emph{Provenance and entity ledger:} Verifies that identifiers, amounts (including European/Swiss formats), unambiguous dates, CJK/Unicode slots, and verbatim note citations trace back to authorized dossier spans before form fills or commits.
  \item \emph{Policy decision table and defeasible branches:} Enforces step-0 missing-parameter pauses, consent rules, access-management checks, reactive checks when dossier fields update, and explicit exception clauses (\emph{defeaters}) before firing a compiled rule.
  \item \emph{Accessibility grammar and compositor check:} Parses DOM actions via AST, pierces open shadow roots and nested frames, checks pointer occlusion, quarantines injected overlays, and flags opaque \texttt{<canvas>} elements for visual hover probing.
  \item \emph{Local taint screening:} Hot-swaps a 12\,MB prompt-injection adapter (\texttt{taint-guard}) onto the shared quantized 4B base model (3.08\,GB VRAM, 13.2\,ms median swap) to block indirect prompt injection on untrusted text.
\end{enumerate}

\paragraph{Level 2: Completion closure, graded UI read-back, and chrome-normalized visual check.}
Before accepting episode completion, \sys{} enforces completion closure: step-limit defaults to deny rather than accept, required evidence tabs are discovered through standard W3C accessibility roles (and verified as read-only via state-involution probes inside an offline twin), and post-commit obligations are tracked across tabs.

Because enterprise apps differ in idempotency and rarely expose direct APIs, \sys{} does not require page reloads or out-of-band database access. Instead, Level~2 uses a graded UI verification ladder:
(i)~\texttt{same\_screen} (universal baseline) inspects the post-save screen transition without reloading, verifying that the form left dirty edit state, no inline validation error appeared, and the saved values or confirmation ID are visible (safe on non-idempotent \texttt{POST} wizards, desktop modals, and remote desktops);
(ii)~\texttt{reopened} (optional when idempotent or when the target lives off-screen) navigates to the target tab, reopens the entity by identity key (filtering hidden inputs, comments, scripts, and substring collisions such as \texttt{\$500} inside \texttt{\$5000}), or parses exported PDF streams while filtering invisible text; and
(iii)~\texttt{separate\_session} (optional) reads the record back from an isolated read-only browser session when stricter separation is required.

When structural DOM signals are absent or ambiguous (such as native desktop apps), Level~2 invokes a quantized 4B local visual verifier (\textsc{Veto-4B}, \texttt{Qwen3.5-4B} NF4 LoRA, 81\,MB adapter, 3.13\,GB VRAM, 118.6\,ms median latency, \$0 API cost) with constrained 3-token decoding (\{\texttt{PASS}, \texttt{VETO}, \texttt{ABSTAIN}\}). Because 94.7\%--95.7\% of pixels in a $384\times 384$ control crop are static chrome that shifts sharply between light and dark themes, \sys{} applies \emph{Dim-the-Chrome} preprocessing: dimming unchanged background pixels by $0.2\times$ and compositing a structured accessibility diff header (49--65\,ms). On a real Chromium portal across light and dark themes, this cuts cross-theme pixel L2 RMSE by 78.5\% (0.8276 $\rightarrow$ 0.1781) and lifts cosine similarity from 0.5385 to 0.8653.

\paragraph{Level 3: System-of-record read-back and signed receipts.}
When an out-of-band channel exists (SQL, FHIR, REST, or filesystem), Level~3 checks before/after row counts and canonical field values. Multi-step episodes record their weakest verification rung (\texttt{checked\_how}) inside an Ed25519-signed in-toto receipt~\cite{torresarias2019intoto}. Unbounded pixel-only checks stay labelled \texttt{VISUAL\_ONLY} and never sign \texttt{VERIFIED}.

\subsection{Regime-Matched Calibration}

Out-of-distribution themes shift raw visual verifier logits and cause false vetoes. Rather than conflating unlabeled normalization with conformal prediction, \sys{} separates three explicit regimes:
\begin{enumerate}
  \item \emph{Zero-label session-start normalization:} Over the first $K_{\text{nav}}$ benign navigation frames in an unseen UI, \sys{} estimates $(\hat{\mu}_0, \hat{\sigma}_0)$ and flags commits only when the normalized anomaly score exceeds $z$. No conformal coverage guarantee is claimed in this regime.
  \item \emph{Commit-Feedback ACI (labeled):} Online Adaptive Conformal Inference~\cite{gibbs2021aci} updates \emph{only} when Level~2 or Level~3 read-back resolves the realized label $Y_t$:
  \begin{equation}
    \alpha_{t+1} = \alpha_t + \gamma\bigl(\alpha - \mathbb{I}[s(x_t, Y_t) > \hat{q}_{1-\alpha_t}]\bigr),
  \end{equation}
  yielding the finite-sample error-rate bound $|\frac{1}{T}\sum_{t=1}^T \text{err}_t - \alpha| \le \frac{1+\gamma}{\gamma T}$ over the labeled commit stream.
  \item \emph{Label-free escalation-rate tracking:} In strictly pixel-only environments without read-back labels $Y_t$ (Section~\ref{sec:rq3}), a Robbins-Monro recursion $\tau_{t+1} = \tau_t + \gamma(\mathbb{I}[s_t > \tau_t] - \rho)$ initialized at $\hat{\mu}_0 + 1.5\hat{\sigma}_0$ stabilizes operator review load near target rate $\rho$ without claiming supervised conformal miscoverage.
\end{enumerate}

\subsection{Crash Recovery and Zero-Egress Self-Healing Compilation}

\paragraph{Write-ahead saga log and bounded repair.}
Long-running batches fail when workers crash (\texttt{SIGKILL}), sessions expire mid-form, or reverse proxies drop responses after the backend already committed. Before each commit sink, \sys{} writes a durable \texttt{PENDING} record keyed by item identity and screen fingerprint to an on-disk write-ahead saga log (WAL). After a crash or re-authentication redirect, the wrapper reconciles pending items through browser-UI read-back (zero backend API calls), skips already-committed items, and resumes from the first uncommitted record. On a read-back mismatch, \sys{} returns the exact expected-versus-observed field diff for up to $R_{\max}=2$ repair turns. On missing parameters or out-of-support policy states, it pauses and emits a single-widget \texttt{PrepareCard} for human sign-off, folding the operator's resolution back into the task cartridge for future runs.

\paragraph{Distilled replay and zero-egress twin repair.}
Verified traces lift into \texttt{ControlSignature}-first compiled plans (amortizing discovery and repair cost across $m$ warm runs as $(C_{\text{discover}} + r C_{\text{heal}})/m \rightarrow 0$). When an upstream app release breaks selectors, \sys{} first re-grounds against an indexed map of live DOM controls. If schema expansion breaks 1-to-1 re-grounding (for example, a new required filter field), \sys{} projects the page into an offline structural twin using bijective session-local symbolization (\texttt{SYM\_*} tokens and order-preserving numeric rank tokens under a strict W3C HTML/ARIA allowlist). A cloud coding model repairs the symbolic script against the twin with 0 bytes of raw payload or patient-data egress, and the repaired plan is promoted only after live read-back passes.

\section{Experiments}
\label{sec:experiments}

We evaluate \sys{} on external public benchmarks, real upstream Docker software upgrades, and real Windows H.264 remote-desktop runs. Headline rates report exact 95\% Wilson confidence intervals, and paired comparisons report exact two-sided McNemar or DeLong tests, all computed directly from raw JSON logs (full artifact inventory in Appendix~\ref{app:repro}). No benchmark tasks, URLs, screenshots, or DOM trees appear in any local adapter training split (Appendix~\ref{app:training}).

\begin{figure*}[t]
\centering
\includegraphics[width=0.85\textwidth]{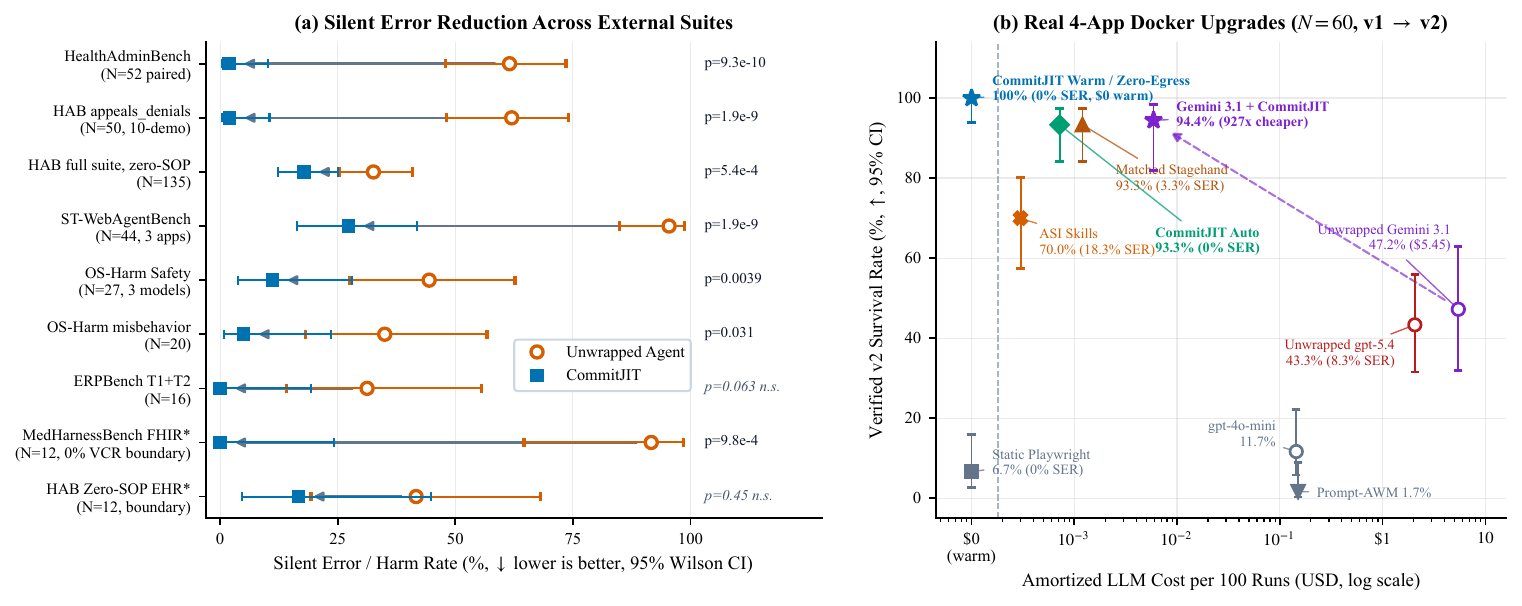}
\caption{\textbf{Overview of external benchmark results.} \textbf{(a)} Paired Silent Error Rate (SER) with 95\% Wilson confidence intervals across external suites, including two non-significant rows (\texttt{ERPBench}, $p=0.0625$; zero-SOP EHR boundary, $p=0.45$) and the \texttt{MedHarnessBench} clinical-judgment boundary (marked $*$, where Arm~3 abstained on 12/12 runs, giving 0\% SER at 0\% VCR). SER alone can be lowered by abstaining; VCR or completion for each row appears in the corresponding table or in Section~\ref{sec:rq2}. \textbf{(b)} Verified v2 survival vs.\ amortized LLM cost across $N=60$ real upstream Docker software upgrades (four apps, v1 $\rightarrow$ v2).}
\label{fig:overview}
\end{figure*}

\subsection{Long-Horizon Web Policy and Clinical Admin Workflows}
\label{sec:rq1}

\paragraph{\texttt{HealthAdminBench}: multi-tab verification, few-shot induction, and the zero-SOP boundary.}
\texttt{HealthAdminBench}~\cite{healthadminbench2026} tests multi-tab EHR, payer-portal, and fax administration workflows. We evaluate four slices under strict clean-room conditions (zero benchmark URLs or task hints, real browser read-back):
(i)~paired frontier ablations ($N=24$ and scaled $N=52$: 26 tasks $\times$ \{\texttt{gpt-5.4}, \texttt{claude-sonnet-5-5}\}, 180 browser reloads);
(ii)~a de-confounded same-model (\texttt{Gemini~4}) full 60/60 longitudinal study on \texttt{appeals\_denials} with pairwise disjoint splits ($\mathcal{D}_{\text{demo}}=10 \rightarrow \mathcal{D}_{\text{stream1}}=20 \rightarrow \mathcal{D}_{\text{holdout2}}=30$);
(iii)~an open-weight diagnostic on \texttt{Qwen3-VL-8B-Instruct} ($N=12$); and
(iv)~a live zero-SOP boundary test on a 12-task EMR portal slice with \texttt{gpt-6.1-terra} using a generic wrapper with zero domain rules.

\begin{table*}[t]
\centering
\caption{\textbf{\texttt{HealthAdminBench} results across scaled paired runs, longitudinal 10-demo induction, and the zero-SOP boundary.} Brackets show 95\% Wilson CIs. Generic wrapping cuts SER sharply when checks apply, but cannot infer unwritten clinic policies at zero demos.}
\label{tab:hab}
\footnotesize
\renewcommand{\arraystretch}{0.95}
\setlength{\tabcolsep}{3.4pt}
\resizebox{\textwidth}{!}{%
\begin{tabular}{llccccc}
\toprule
\textbf{Slice / Model} & \textbf{Configuration} & \textbf{Full Comp.} & \textbf{Subtask} & \textbf{VCR ($\uparrow$)} & \textbf{SER ($\downarrow$, 95\% CI)} & \textbf{HEP} \\
\midrule
Core Frontier ($N=24$) & Unwrapped $\rightarrow$ Single-Tab Reload & 29.2\% $\rightarrow$ 41.7\% & 68.3\% $\rightarrow$ 72.3\% & 29.2\% $\rightarrow$ 20.8\% & 70.8\% $\rightarrow$ 45.8\%$^\dagger$ & 37.5\% (3/8) \\
Core Frontier ($N=24$) & Unwrapped $\rightarrow$ Multi-Tab Ledger & 41.7\% $\rightarrow$ 50.0\% & 75.0\% $\rightarrow$ 71.7\% & 41.7\% $\rightarrow$ 45.8\% & 58.3\% $\rightarrow$ 41.7\% & 66.7\% (2/3) \\
\midrule
Scaled Frontier ($N=52$) & Unwrapped (26 tasks $\times$ 2 models) & 38.5\% (20/52) & 73.4\% & 38.5\% (20/52) & 61.5\% [48.0, 73.5] & n/a \\
Scaled Frontier ($N=52$) & \sys{} (+ Multi-Tab \& Default-Deny) & 44.2\% (23/52) & 74.3\% & 11.5\% (6/52) & \textbf{1.9\% [0.3, 10.1]}$^\ddagger$ & 62.2\% (28/45) \\
\midrule
Appeals Held-Out ($N=50$) & \texttt{Gemini~4} Unwrapped (0 demos) & 38.0\% (19/50) & 85.0\% & 38.0\% (19/50) & 62.0\% [48.2, 74.1] & n/a \\
Appeals Held-Out ($N=50$) & \sys{} 10-Demo Cartridge (0.00 touches) & 20.0\% (10/50) & 50.9\% & 20.0\% (10/50) & \textbf{2.0\% [0.4, 10.5]}$^\S$ & 90.9\% prec.$^\P$ \\
Appeals Held-Out ($N=50$) & \sys{} Longitudinal (+ 1-Slot \texttt{PrepareCard}) & 38.0\% (19/50) & 70.4\% & 32.0\% (16/50) & \textbf{10.0\% [4.3, 21.4]}$^\S$ & 0.32 touch/ep \\
\midrule
Zero-SOP EHR ($N=12$)$^*$ & \texttt{gpt-6.1-terra} Unwrapped $\rightarrow$ Generic Wrap & 16.7\% $\rightarrow$ 0.0\% & --- & 16.7\% $\rightarrow$ 0.0\% & 41.7\% $\rightarrow$ 16.7\% ($p=0.45$) & 16.7\% \\
Open-Weight ($N=12$) & \texttt{Qwen3-VL-8B} Unwrapped $\rightarrow$ \sys{} & 0.0\% $\rightarrow$ 0.0\% & 14.4\% $\rightarrow$ 23.9\% & 0.0\% $\rightarrow$ 0.0\% & 100\% $\rightarrow$ 75.0\% & 100\% (3/3) \\
\bottomrule
\multicolumn{7}{l}{\footnotesize $^\dagger$$p=0.031$. $^\ddagger$31 vs.\ 0 ($p=9.31\times 10^{-10}$; \texttt{claude-sonnet-5-5} reached 0.0\% SER, $p=1.22\times 10^{-4}$). $^\S$0-touch $p=1.86\times 10^{-9}$; longitudinal $p=2.98\times 10^{-8}$.}\\
\multicolumn{7}{l}{\footnotesize $^\P$Commit precision 10/11 (90.9\%); 39/50 tasks paused for \texttt{PREPARE}. $^*$Live EMR portal with zero domain policy specs (boundary result).}
\end{tabular}%
}
\end{table*}

Table~\ref{tab:hab} separates what commit-boundary verification accomplishes from where it trades off coverage:
\begin{itemize}
  \item \emph{Single-tab vs.\ multi-tab read-back ($N=24$):} Reloading a single tab cut SER from 70.8\% to 45.8\% ($p=0.031$) but falsely escalated 5 passing multi-tab runs, dropping VCR from 29.2\% to 20.8\% (37.5\% HEP). Adding the cross-tab entity ledger recovered VCR to 45.8\% (+25.0\,pp over single-tab reload) at 66.7\% HEP.
  \item \emph{Safety vs.\ autonomous coverage ($N=52$):} Default-deny closure cut SER from 61.5\% (32/52) to 1.9\% (1/52; 31 vs.\ 0, exact McNemar $p=9.31\times 10^{-10}$) while lifting raw completion to 44.2\%. However, strict default-deny escalated 45/52 runs (62.2\% HEP), reducing autonomous VCR from 38.5\% to 11.5\%.
  \item \emph{Few-shot policy induction vs.\ zero-SOP boundary:} On $\mathcal{D}_{\text{stream1}}$ ($N=20$), zero-guideline gating remained at 65.0\% (13/20) SER, and on the live 12-task zero-SOP EHR test with \texttt{gpt-6.1-terra}, generic wrapping produced 0.0\% VCR vs.\ 16.7\% unwrapped (SER 41.7\% $\rightarrow$ 16.7\%, $p=0.45$, not significant). Wrapping cannot invent clinic policies it was never given. Once 10 raw demonstrations induce a zero-egress cartridge (3/3 HTTP~403 egress canary blocks), held-out SER across all $N=50$ disjoint appeals tasks drops 31$\times$ from 62.0\% (31/50) to 2.0\% (1/50, $p=1.86\times 10^{-9}$, 90.9\% commit precision). Adding simulated 1-slot \texttt{PrepareCard} sign-off on $\mathcal{D}_{\text{stream1}}$ (0.80 touches/episode, 0.32 pooled) recovers VCR to 50.0\% on $\mathcal{D}_{\text{stream1}}$ (32.0\% pooled) at 10.0\% pooled SER ($p=2.98\times 10^{-8}$).
\end{itemize}

\paragraph{\texttt{ST-WebAgentBench}: step critic vs.\ commit-boundary gating across 3 enterprise apps.}
We evaluate official \texttt{ST-WebAgentBench}~\cite{levy2024stwebagentbench} in three scopes: a 3-way step-critic ablation on \texttt{ShoppingAdmin} ($n=8$), a paired 3-app suite ($N=44$: \texttt{ShoppingAdmin} $n=8$, \texttt{GitLab} $n=18$, \texttt{SuiteCRM} $n=18$), and an expanded 3-app suite ($N=100$; Table~\ref{tab:stweb}).

\begin{table}[t]
\centering
\caption{\textbf{\texttt{ST-WebAgentBench} across \texttt{ShoppingAdmin} ($n=8$), paired 3-app ($N=44$), and expanded 3-app ($N=100$).}}
\label{tab:stweb}
\footnotesize
\setlength{\tabcolsep}{2.2pt}
\resizebox{\columnwidth}{!}{%
\begin{tabular}{lccccc}
\toprule
\textbf{Scope / Variant} & \textbf{CR / CuP} & \textbf{SER ($\downarrow$, 95\% CI)} & \textbf{HEP} & \textbf{Crit./ep} & \textbf{$p$} \\
\midrule
ShopAdmin Unwrapped ($n{=}8$) & 0.0 / 0.0\% & 100.0\% [67.6, 100] & n/a & 0.00 & ref. \\
ShopAdmin Step Critic ($n{=}8$) & 0.0 / 0.0\% & 37.5\% [13.7, 69.4] & 100\% (5/5) & 7.38 & 0.0625 \\
ShopAdmin Commit-Only ($n{=}8$) & 0.0 / 0.0\% & \textbf{12.5\%} [2.2, 47.1] & 100\% (7/7) & \textbf{1.38} & 0.0156 \\
\midrule
3-App Unwrapped ($N{=}44$) & 6.8 / 4.5\% & 95.5\% [84.9, 98.7] & n/a & 0.00 & ref. \\
3-App Commit-Only ($N{=}44$) & 11.4 / 6.8\% & 50.0\% [35.8, 64.2] & 100\% (19/19) & 1.57 & $2.0{\times}10^{-6}$ \\
3-App + Policy \& \texttt{PREPARE} & \textbf{13.6 / 6.8\%} & \textbf{27.3\%} [16.4, 41.9] & \textbf{100\% (29/29)} & 2.80 & $1.9{\times}10^{-9}$ \\
\midrule
Sub.\ Agent Paired ($N{=}44$) & 13.6 / 4.5\% & 100 $\rightarrow$ \textbf{27.3\%} & \textbf{100\% (30/30)} & 2.23 & $4.7{\times}10^{-10}$ \\
Expanded 3-App ($N{=}100$) & \textbf{15.0 / 7.0\%} & \textbf{25.0\%} [17.6, 34.3] & \textbf{98.6\% (68/69)} & 2.01 & --- \\
\bottomrule
\end{tabular}%
}
\end{table}

On \texttt{ShoppingAdmin}, a step-level critic consumes 59 calls (7.38/episode) and leaves 37.5\% SER; commit-only gating cuts SER to 12.5\% in 5.4$\times$ fewer calls (11 vs.\ 59, $p=0.0156$). On paired 3-app $N=44$ (\texttt{gpt-5.4}), adding before-step policy checks and \texttt{PREPARE} cuts SER from 95.5\% (42/44) to 27.3\% (12/44; 30 vs.\ 0, $p=1.86\times 10^{-9}$) at 100\% HEP (29/29), while doubling raw completion from 6.8\% to 13.6\% at matched token usage (2.09M vs.\ 2.09M tokens). Across $N=100$ live episodes on a subscription CLI agent, SER stays at 25.0\% (25/100) with 98.6\% (68/69) escalation precision.

\subsection{Calibration, Prompt Injection, and Safety Tradeoffs}
\label{sec:rq2}

\paragraph{\texttt{AgentDojo} (512 deterministic + 128 calibrated Multi-LoRA episodes).}
Across all four \texttt{AgentDojo v1.2.1} suites~\cite{debenedetti2024agentdojo} (512 episodes, 4 models), deterministic before-commit checks cut attack success rate (ASR) on \texttt{gemini-2.5-flash} from 72.9\% to 16.7\% while raising benign utility from 37.5\% to 56.2\%, cut \texttt{Qwen3.5-4B} ASR from 14.6\% to 2.1\%, and caused 0 false vetoes on \texttt{gpt-5.4} (benign utility 81.2\% $\rightarrow$ 87.5\%, 0.0\% ASR) and \texttt{claude-sonnet-5-5} (100.0\% benign utility, 0.0\% ASR).

When mounting \texttt{taint-guard} concurrently with \textsc{Veto-4B-r5} on Slack and Travel ($N=128$: 96 attack, 32 benign), a static split-conformal threshold ($\tau_{\text{cal}} = -10.25$) drove ASR to 0/96 (0.0\%, $p=1.19\times 10^{-7}$), but 31 OOD false vetoes dropped benign utility from 78.1\% to 53.1\% and under-attack utility from 75.0\% to 45.8\%. Switching to sink-only gating with OOD-safe calibration ($\tau = 0.0$) recovered +34.4\,pp on both metrics (87.5\% benign utility, 80.2\% under-attack utility) while keeping pooled ASR at 3.1\% (3/96; 0/48 on Travel).

\paragraph{Stanford EHR calibration ($N=72$ full-screen, $N=82$ crop).}
On unseen EHR audit streams, uncalibrated \textsc{Veto-4B-r5} suffered a 98.5\% (66/67) full-screen false-veto rate (81.7\% on crops). Zero-label session-start $z$-score normalization ($K=3, z>2.0$ or $K=5, z>1.5$) cut false vetoes to 0.0\% (0/64 full-screen, 0/77 crop). Label-free Robbins-Monro tracking ($\rho=0.05, \gamma=0.15$) held false vetoes at 1.6\% (1/64). Supervised Commit-Feedback ACI ($\alpha=0.05, \gamma=0.15$, updated strictly on realized read-back $Y_t$) achieved 0.0\% (0/64) in threshold form ($K=3$) and 6.8\% (4/59) in exact quantile form ($K=8$, within the theoretical bound $\frac{1+\gamma}{\gamma T} = 13.0\%$), catching 5/5 injected two-plane faults.

\paragraph{\texttt{OS-Harm} and \texttt{AgentCIBench} (including utility tradeoff).}
On official \texttt{OS-Harm}~\cite{osharm2025} ($N=54$ isolated Docker runs across 9 tasks and 3 models), \sys{} cut aggregate harm by 75\%, from 44.4\% (12/27, [27.6\%, 62.7\%]) to 11.1\% (3/27, [3.9\%, 28.1\%], exact McNemar $p=0.00391$), lifting policy-compliant completion from 40.7\% to 63.0\% ($p=0.070$) at nearly identical API cost (\$0.0387 vs.\ \$0.0381/task) and 0/24 harm on all 8 policy-covered tasks. On a separate 20-task \texttt{OS-Harm} Docker misbehavior slice with \texttt{gpt-6.2-sol} and copy-on-write filesystem rollback, SER dropped from 35.0\% (7/20) to 5.0\% (1/20; exact McNemar $p=0.03125$) while VCR stayed at 60.0\% (12/20) in both arms ($p=1.0$), with 7 rollbacks restored. Version~1 reported an earlier wrapper on the same 20 tasks (SER 40.0\% $\rightarrow$ 0.0\%, $p=0.0078$) whose conservative completion checks cut VCR from 55.0\% to 20.0\%; the revised wrapper removes most of that coverage loss.

On \texttt{AgentCIBench} ($N=25$ end-to-end contextual-integrity tasks, \texttt{gpt-6.2-sol}), generic egress gating lowered privacy leak rate from 12.0\% (3/25) to 4.0\% (1/25), at the cost of lowering task utility from 80.0\% to 64.0\% ($p=0.50$, not significant). On \texttt{BrowserGym MiniWoB++} overlay injection ($N=80$, Appendix~\ref{app:enterprise}), AST accessibility decoding raised pooled adversarial completion from 45.0\% to 90.0\% ($p=8.0\times 10^{-6}$, 0.0\% policy violations).

\subsection{Pixel-Only Desktop and Remote-Desktop Boundaries}
\label{sec:rq3}

To test Propositions~\ref{prop:compounding} and~\ref{prop:unidentifiable} where no DOM exists, we evaluate \textsc{Veto-4B} on two pure-pixel surfaces: Ubuntu desktop trajectories from \texttt{OSWorld-Verified} and live Windows H.264/AVC444 remote-desktop streams.

\begin{table}[t]
\centering
\caption{\textbf{Pixel-only \texttt{OSWorld-Verified} ($N=120$ Opus~4.5 desktop runs, 10 Ubuntu apps).} Confirms Proposition~\ref{prop:compounding} (step-level 100\% false vetoes vs.\ 17.5--19.0\% calibrated commit-sink) and Proposition~\ref{prop:unidentifiable} (0.523 1-frame visual AUROC).}
\label{tab:osworld}
\footnotesize
\setlength{\tabcolsep}{2.5pt}
\resizebox{\columnwidth}{!}{%
\begin{tabular}{lcccc}
\toprule
\textbf{Variant} & \textbf{Fail Recall ($\uparrow$)} & \textbf{False Vetoes ($\downarrow$)} & \textbf{Benign Util.} & \textbf{p50 ms} \\
\midrule
Unwrapped & 0.0\% (0/57) & 0.0\% (0/63) & 100.0\% & 0 \\
Step-Level VLM & 100.0\% & 100.0\% (63/63) & 0.0\% & 2,945 \\
Commit Raw 4B & 35.1\% (20/57) & 38.1\% (24/63) & 61.9\% & 517 \\
Dim + Calib ($z>1.5$) & 29.8\% (17/57) & 20.6\% (13/63) & 79.4\% & 487 \\
Dim + Calib ($z>2.0$) & 26.3\% (15/57) & \textbf{17.5\% (11/63)} & \textbf{82.5\%} & \textbf{487} \\
Dim + Online Thresh. & 28.1\% (16/57) & 19.0\% (12/63) & 81.0\% & \textbf{487} \\
\bottomrule
\end{tabular}%
}
\end{table}

\paragraph{\texttt{OSWorld-Verified}: 1-frame compounding ($N=120$) and strict 4-app unseen holdout ($N=160$).}
On $N=120$ real \texttt{claude-opus-4.5} trajectories across 10 Ubuntu apps (63 clean, 57 silent failures; Table~\ref{tab:osworld}), step-level visual gating compounded to 100.0\% (63/63, [94.3\%, 100.0\%]) false vetoes and 0.0\% benign utility at 2,945\,ms p50. Moving \textsc{Veto-4B-r5} to commit sinks with Dim-the-Chrome and session-start normalization or label-free escalation tracking ($\rho=0.10$) preserved 81.0\%--82.5\% benign utility at 487\,ms (6.0$\times$ faster, McNemar $p < 10^{-15}$ vs.\ step-level; $p = 0.029$ vs.\ raw commit 4B). Zero-shot single-frame AUROC remained 0.523.

Training a 2-frame $[\text{Dim}(\text{step}_0) \mid \text{Dim}(\text{step}_T)]$ adapter (\textsc{Veto-4B-r6}, 3.15\,GB VRAM, 357\,ms p50) on 6 Ubuntu apps ($N=440$ train, $N_{\text{val}}=9$) and testing on 4 strictly unseen apps (\texttt{gimp}, \texttt{libreoffice\_calc}, \texttt{thunderbird}, \texttt{vlc}; $N=160$: 73 failures, 87 clean) lifted AUROC from 0.6272 to 0.7905 (+0.1633, paired DeLong $p=0.0051$), led by \texttt{gimp} (0.5038 $\rightarrow$ 0.9087) and \texttt{libreoffice\_calc} (0.7963 $\rightarrow$ 0.8625). Session-start normalization lowered false vetoes from 37.9\% (33/87) to 6.9\% (6/87; 93.1\% benign utility, 52.1\% failure recall; 3.4\% FPR at $\tau=1.0$). Consistent with Proposition~\ref{prop:unidentifiable}, the remaining $\sim$45\% of silent failures were non-visual filesystem, mail-folder, or configuration errors that pixels alone cannot detect.

\paragraph{Real Windows H.264/AVC444 RDP streams ($N=105$).}
To measure pixel verification under real video-codec noise, we evaluated 105 counted runs through stock Windows \texttt{mstsc} against Windows Server 2022 with H.264/AVC444 encoding (Patient Notes and OpenEMR, 0 model calls, verified out-of-band against each app's database). Across all 105 runs, \sys{} committed 0/105 wrong-record writes (95\% Wilson upper bound 3.5\%) and its database check agreed with ground truth on 50/50 judged runs. However, it triggered 40/105 false stops (including 12/20 healthy OpenEMR runs), driven by 1--2\,px hover/codec shifts (10/40), unsettled carets or progressive AVC444 refinement (13/40), baked-in row auto-selection (10/40), and layout reflow breaking absolute postcondition regions (5/40; detailed breakdown in Appendix~\ref{app:enterprise}). Pixel-only guards achieve zero wrong saves on remote desktop, but reliable unattended throughput on compressed VDI still requires relative/constellation grounding and settled-frame tolerance.

\subsection{Real Software Upgrades and Crash Recovery}
\label{sec:rq4}

\paragraph{Four-app longitudinal Docker upgrades ($N=60$, v1 $\rightarrow$ v2).}
We benchmark 60 workflows (15 per app) across four pinned upstream Docker release pairs: \texttt{Gitea} \texttt{1.21.11} $\rightarrow$ \texttt{1.23.6}, \texttt{Vikunja} \texttt{0.23.0} $\rightarrow$ \texttt{0.24.6}, \texttt{Memos} \texttt{0.18.2} $\rightarrow$ \texttt{0.24.0}, and \texttt{Mealie} \texttt{v1.12.0} $\rightarrow$ \texttt{v2.8.0}. Ground truth uses independent REST API read-back on v2 (Table~\ref{tab:ui_drift}).

\begin{table*}[t]
\centering
\caption{\textbf{Four-app upstream Docker software upgrades ($N=60$, v1 $\rightarrow$ v2; REST API ground truth).} \sys{} reaches 93.3\% autonomous and 100.0\% zero-egress verified survival at 0.0\% SER and \$0.00 warm replay, improving every frontier model tested.}
\label{tab:ui_drift}
\footnotesize
\renewcommand{\arraystretch}{0.92}
\setlength{\tabcolsep}{3.6pt}
\resizebox{\textwidth}{!}{%
\begin{tabular}{lccccc}
\toprule
\textbf{Method / Model} & \textbf{v2 Verified (95\% CI)} & \textbf{Claimed} & \textbf{SER ($\downarrow$)} & \textbf{p50 ms} & \textbf{Cost / 100 runs} \\
\midrule
Static Playwright ($N=60$) & 6.7\% (4/60) [2.6, 15.9] & 4/60 & 0.0\% & 2,825 & \$0.0000 \\
Prompt-AWM~\cite{wang2025awm} ($N=60$) & 1.7\% (1/60) [0.3, 8.9] & 0/60 & 0.0\% & 10,710 & \$0.1505 \\
Unwrapped \texttt{gpt-4o-mini} ($N=60$) & 11.7\% (7/60) [5.8, 22.2] & 0/60 & 0.0\% & 11,330 & \$0.1444 \\
ASI Programmatic Skills~\cite{wang2025asi} ($N=60$) & 70.0\% (42/60) [57.5, 80.1] & 53/60 & 18.3\% (11/60) & 1,882 & \$0.0003 \\
\sys{} Selector-Only ($N=60$) & 70.0\% (42/60) [57.5, 80.1] & 42/60 & \textbf{0.0\% (0/60)} & 1,884 & \$0.0015 \\
Stagehand Unverified (Matched \texttt{gpt-5.4}, $N=60$) & 93.3\% (56/60) [84.1, 97.4] & 58/60 & 3.3\% (2/60) & 1,910 & \$0.0009 \\
\sys{} + Indexed Live-DOM (Autonomous, $N=60$) & 93.3\% (56/60) [84.1, 97.4] & 56/60 & \textbf{0.0\% (0/60)} & 1,890 & \$0.0009 \\
\sys{} + Zero-Egress Twin Repair ($N=60$, 0\,B raw) & \textbf{100.0\% (60/60)} [94.0, 100] & 60/60 & \textbf{0.0\% (0/60)} & 1,782 & \$0.0011 \\
\sys{} + Cold$\rightarrow$Warm \texttt{PREPARE} ($N=60$, 2 cards) & \textbf{100.0\% (60/60)} [94.0, 100] & 60/60 & \textbf{0.0\% (0/60)} & 1,843 & \textbf{\$0.0000 (warm)} \\
\midrule
Unwrapped \texttt{gpt-5.4} ($N=60$) & 43.3\% (26/60) [31.6, 55.9] & 31/60 & 8.3\% (5/60) & 10,575 & \$2.0579 \\
Unwrapped \texttt{gemini-3.1-pro-preview} ($N=36$)$^\dagger$ & 47.2\% (17/36) [32.0, 63.0] & 17/36 & 0.0\% (0/36) & 34,087 & \$5.4532 \\
+ \sys{} (\texttt{gemini-3.1-pro-preview}, $N=36$)$^\dagger$ & \textbf{94.4\% (34/36)} [81.9, 98.5] & 34/36 & \textbf{0.0\% (0/36)} & 2,417 (warm) & \textbf{\$0.0059} \\
Unwrapped \texttt{Gemini~4} ($N=60$, 230 calls) & 93.3\% (56/60) [84.1, 97.4] & 58/60 & 3.3\% (2/60) & 21,776 & 230 calls \\
+ \sys{} (\texttt{Gemini~4}, $N=60$, 53 calls; 47 warm) & \textbf{100.0\% (60/60)} [94.0, 100] & 60/60 & \textbf{0.0\% (0/60)} & 1,864 (warm) & 53 calls (0 warm) \\
Unwrapped \texttt{Claude~5.5} ($N=60$, 215 calls) & 95.0\% (57/60) [86.3, 98.3] & 59/60 & 3.3\% (2/60) & 39,613 & 215 calls \\
+ \sys{} (\texttt{Claude~5.5}, $N=60$, 72 calls; 45 warm) & \textbf{100.0\% (60/60)} [94.0, 100] & 60/60 & \textbf{0.0\% (0/60)} & 1,959 (warm) & 72 calls (0 warm) \\
\bottomrule
\multicolumn{6}{l}{\footnotesize $^\dagger$Paired $N=36$ before Tier-1 250 RPD API quota ($p=1.53\times 10^{-5}$, 927$\times$ lower cost). Pooled \texttt{Gemini~4} + \texttt{Claude~5.5} ($N=120$): 7 vs.\ 0 ($p=0.0156$).}
\end{tabular}%
}
\end{table*}

Indexed live-DOM re-grounding lifts autonomous \sys{} from 70.0\% to 93.3\% (56/60; $p=0.000122$) with 0.0\% SER, whereas unverified ASI~\cite{wang2025asi} incurs 18.3\% (11/60) SER ($p=0.000977$) and matched \texttt{gpt-5.4} Stagehand incurs 3.3\% (2/60) SER (e.g., \texttt{Vikunja} silently stripping date words from titles). Zero-egress symbolic twin repair resolves all 4 remaining schema-expansion tasks (60/60, 0\,B raw egress, 3/3 egress red-team blocks); equivalently, 2/60 \texttt{PrepareCard}s reach 60/60 warm replay at 1.84\,s and \$0.00 while catching 20/20 simulated tired-human errors. Across multi-vendor frontier models, \sys{} lifts \texttt{gemini-3.1-pro-preview} from 47.2\% to 94.4\% ($p=1.53\times 10^{-5}$, 927$\times$ cheaper), \texttt{Gemini~4} from 93.3\% to 100.0\% (11.7$\times$ faster warm), and \texttt{Claude~5.5} from 95.0\% to 100.0\% (20.2$\times$ faster warm; pooled frontier $N=120$: 7 vs.\ 0, $p=0.0156$).

\begin{figure*}[t]
\centering
\includegraphics[width=0.79\textwidth]{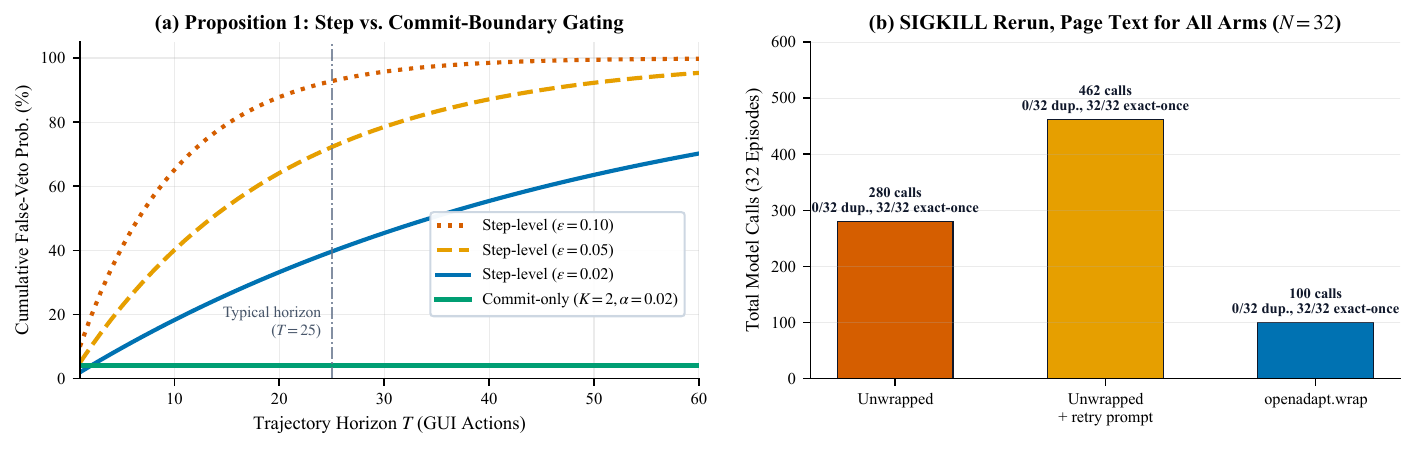}
\caption{\textbf{Compounding theory and the fair crash rerun.} \textbf{(a)} Cumulative false-veto probability under Proposition~\ref{prop:compounding} as horizon $T$ grows. \textbf{(b)} Mid-batch \texttt{SIGKILL} rerun ($N=32$: 16 original and 16 replication tasks, \texttt{gpt-6.2-sol}, 0 backend API calls) in which every arm sees the URL, page title, visible page text, and interactive elements and keeps a notes field across steps. All three arms finish 32/32 episodes exactly once with 0/32 duplicates; \sys{} uses 100 model calls against 280 unwrapped and 462 with a retry prompt.}
\label{fig:compounding}
\end{figure*}

\paragraph{\texttt{EnterpriseSessionBench}, \texttt{DesktopChaosBench}, and open evaluation harness.}
We release our 4-arm evaluation harness (reporting VCR alongside SER) and custom fault suites in \ifarxiv\texttt{openadapt-evals}\else our open evaluation repository\fi, together with adapters for \texttt{ERPBench}, \texttt{MedHarnessBench}, \texttt{WindowsWorld}, \texttt{SCUBA}, \texttt{WoW-bench}, and \texttt{Continual GUI}. On \texttt{EnterpriseSessionBench}, we evaluate exact-once execution across the same four Docker apps using browser-UI read-back only (0 backend API calls during execution) under four failure classes: session expiry redirects (Fault A), step-up MFA walls (Fault B), post-commit dropped responses (Fault C), and mid-batch \texttt{SIGKILL} worker crashes that wipe agent memory (Fault D; Table~\ref{tab:esb} and Figure~\ref{fig:compounding}(b)).

\begin{table*}[t]
\centering
\caption{\textbf{\texttt{EnterpriseSessionBench}: live Docker fault injection (0 backend API calls).} With page text visible to every arm, the worker-crash suite shows no duplicate or silent-error gap; the remaining difference is model calls. The auth-fault rows keep v1's controls-only baseline.}
\label{tab:esb}
\footnotesize
\renewcommand{\arraystretch}{0.92}
\setlength{\tabcolsep}{3.4pt}
\begin{adjustbox}{max width=\textwidth}
\begin{tabular}{llccccc}
\toprule
\textbf{Suite / Faults} & \textbf{Arm (Model)} & \textbf{VCR ($\uparrow$)} & \textbf{Dup.\ Writes ($\downarrow$)} & \textbf{SER ($\downarrow$)} & \textbf{LLM Calls} & \textbf{p50 s} \\
\midrule
Auth-Chaos ($N=48$, 72 rec.)$^*$ & Unwrapped (\texttt{Gemini~4}, controls-only) & 83.3\% (40/48) & 18.8\% (9/48) & 16.7\% (8/48) & 359 & 56.0 \\
Faults A/B/C (not rerun) & \sys{} (\texttt{Gemini~4}) & 100.0\% (48/48)$^\dagger$ & 0.0\% (0/48) & 0.0\% (0/48) & 206 (-42.6\%) & 27.2 \\
\midrule
\texttt{SIGKILL} rerun ($N=32$, 64 rec.) & Unwrapped (\texttt{gpt-6.2-sol}) & 100.0\% (32/32) & 0.0\% (0/32) & 0.0\% (0/32) & 280 & n/r \\
Fault D after item 1, & Unwrapped + retry prompt & 100.0\% (32/32) & 0.0\% (0/32) & 0.0\% (0/32) & 462 (+65.0\%) & n/r \\
page text for all arms & \sys{} \texttt{iid\_ephemeral} & 100.0\% (32/32)$^\ddagger$ & 0.0\% (0/32) & 0.0\% (0/32) & \textbf{100 (-64.3\%)} & n/r \\
\bottomrule
\end{tabular}
\end{adjustbox}
\par\vspace{1pt}{\footnotesize\raggedright $^*$Unwrapped agent saw only interactive elements, no page text; not rerun. All 9 duplicate writes came from post-commit dropped responses (Fault C). $^\dagger$Exact-once $p=0.00781$, duplicates $p=0.00391$. $^\ddagger$No discordant pairs: every paired exact McNemar $p=1.0$ (VCR, duplicates, SER). 0/32 is consistent with a true duplicate rate up to about 9\% (one-sided exact 95\% bound). n/r: not recorded.\par}
\end{table*}

\paragraph{Erratum: the v1 crash-duplicate gap came from a blind baseline.}
Version~1 reported that after a \texttt{SIGKILL} following item~1, unwrapped agents duplicated a record in 10 of 16 episodes (\texttt{gpt-6.2-sol}) and 11 of 16 (\texttt{gpt-5.6-sol}) against 0 of 16 wrapped, and pooled these with Faults A/B/C into a 64-episode duplicate rate of 29.7\% $\rightarrow$ 0.0\%. In those runs the unwrapped agent saw only form controls and links (at most 55) and no page text, and kept an action log but no memory of what it had seen, so it often could not see the record it had already saved. Appending a retry prompt (search for each reference code before creating it) to the same weak view cut duplicates to 5 of 16 in each generation, mostly by turning them into lost writes and unfinished episodes. We then pre-registered a rerun in which all three arms receive the same observation (URL, title, visible page text as collapsed \texttt{innerText}, and the same interactive elements), a notes field carried across steps and wiped by the crash in every arm, and a 20+20 step budget, on the original 16 tasks plus 16 new replication tasks frozen before any run. All three arms reached 32/32 exact-once with 0/32 duplicates and 0 lost items (Table~\ref{tab:esb}). In each unwrapped arm, all 32 restarted workers had their saved record in the page text before their next save (20 on the first frame after the restart, 12 after opening a list or search page). The remaining measured difference is cost: \sys{} used 100 model calls against 280 unwrapped and 462 with the retry prompt, because its restarted worker made no model calls (read-back confirmed item~1 and item~2 replayed from the item-1 trace). The wrapped arm also kept aids the unwrapped arms lack (restoring the form route, a list of verified items, read-back-gated completion). Six inputs changed at once between v1 and the rerun, so we cannot say which one closes the gap.

The lesson for evaluation is that a baseline must see what a careful operator would see; otherwise the wrapper is credited for the baseline's blindness. The rerun covers one worker, a unique reference code per item, a crash right after a confirmed save, and an app that lists the saved record as plain text. Three cases remain where the outcome is not on any screen the agent can read, even for a frontier model: (i)~delayed commits, where the write lands after the screen is read or its acknowledgement is lost; (ii)~concurrent workers acting on the same job; and (iii)~non-searchable writes that never appear in a list or search view. These are the cases that still need a durable write-ahead log and read-back by identity key. Our suite tests only (i), through Fault C, and only against the controls-only baseline, so the Auth-Chaos duplicate gap in Table~\ref{tab:esb} needs the same page-text rerun before it can be taken as a wrapper effect.

On our companion \texttt{DesktopChaosBench} v0.1 suite ($N=25$ rendered $640\times 400$ single-button canvas tasks spanning mixed-DPI letterbox, lossy streams, stripped accessibility trees, focus-steal modals, and RDP disconnects), upgrading from \texttt{gpt-5.6-luna} to \texttt{gpt-6.1-terra} raised unwrapped single-screen accuracy from 12.0\% (3/25, 90.0\,s p50) to 100.0\% (25/25, 13.7\,s p50), matching wrapped 100.0\% at 0.0\% SER. Here a newer model closed the gap; in the crash rerun the model stayed the same and the gap closed once the baseline could read the page. On our \texttt{ExtraDup} wrong-record and duplicate suite ($N=28$, \texttt{gpt-5.6-luna}), wrapping lifted exact-once VCR from 85.7\% (24/28) to 100.0\% (28/28, $p=0.125$, not significant; 0.0\% wrapped SER).

\subsection{Paired Live Runs on Additional External Benchmarks}
\label{sec:wave2}

Table~\ref{tab:wave2} reports six paired live runs (same tasks, same agent model in both arms, \$0 paid API spend through subscription CLIs). Arm~0 is the unwrapped agent, whose receipt reads \texttt{VERIFIED} whenever it declares completion; Arm~3 wraps the same agent with \texttt{openadapt.wrap}. Ground truth is each benchmark's own grader run after the episode. Each benchmark's grader decides pass or fail; VCR counts runs that are \texttt{VERIFIED} and pass, and SER counts runs that are \texttt{VERIFIED} and fail.

\begin{table*}[t]
\centering
\caption{\textbf{Paired live runs on external benchmarks (Arm~0 unwrapped vs.\ Arm~3 \texttt{openadapt.wrap}).} Exact two-sided McNemar $p$ on paired tasks. SER falls or stays at zero in every suite; VCR rises on \texttt{ERPBench}, holds on three suites, and falls where the wrapper abstains on clinical work it cannot check.}
\label{tab:wave2}
\footnotesize
\renewcommand{\arraystretch}{0.95}
\setlength{\tabcolsep}{3.0pt}
\begin{adjustbox}{max width=\textwidth}
\begin{tabular}{llcccccl}
\toprule
& & \multicolumn{3}{c}{\textbf{VCR ($\uparrow$)}} & \multicolumn{3}{c}{\textbf{SER ($\downarrow$)}} \\
\cmidrule(lr){3-5}\cmidrule(lr){6-8}
\textbf{Benchmark ($N$)} & \textbf{Agent} & \textbf{Arm 0} & \textbf{Arm 3} & \textbf{$p$} & \textbf{Arm 0} & \textbf{Arm 3} & \textbf{$p$; other paired measures} \\
\midrule
\texttt{ERPBench} T1+T2 (16) & \texttt{gpt-6.2-sol} & 68.8\% (11) & \textbf{81.3\% (13)} & 0.5 & 31.3\% (5) & \textbf{0.0\% (0)} & 0.0625; Arm 3 receipts match grader 16/16 \\
\texttt{HealthAdminBench} full (135) & \texttt{gpt-6.1-terra} & 6.7\% (9) & 3.7\% (5) & 0.22 & 32.6\% (44) & \textbf{17.8\% (24)} & $5.4{\times}10^{-4}$; zero-SOP; JMESPath check mean 0.483 $\rightarrow$ 0.502 \\
\texttt{OS-Harm} misbehavior (20) & \texttt{gpt-6.2-sol} & 60.0\% (12) & 60.0\% (12) & 1.0 & 35.0\% (7) & \textbf{5.0\% (1)} & 0.03125; 7 rollbacks restored \\
WorkArena L1 (12) & \texttt{gpt-6.2-sol} & 66.7\% (8) & 66.7\% (8) & 1.0 & 0.0\% (0) & 0.0\% (0) & 1.0; dup.\ commits 1 $\rightarrow$ 0; p50 $-$40\% \\
\texttt{MedHarnessBench} FHIR (12) & Claude Sonnet & 8.3\% (1) & 0.0\% (0) & 1.0 & 91.7\% (11) & \textbf{0.0\% (0)} & $9.8{\times}10^{-4}$; Arm 3 abstains on 12/12 \\
Re-hosted WAA, Win~11 (11) & \texttt{gpt-6.2-sol} & 72.7\% (8) & 72.7\% (8)$^*$ & 1.0 & 0.0\% (0) & 0.0\% (0) & 1.0; 31 vs.\ 60 judge calls (prior wrapper) \\
\bottomrule
\end{tabular}
\end{adjustbox}
\par\vspace{1pt}{\footnotesize\raggedright Counts in parentheses. Ground truth: ERPNext DB grader, official \texttt{HealthAdminBench} scorer, per-task safety/success checks, WorkArena \texttt{validate()}, outcome checkpoints, re-implemented WAA checkers. $^*$63.6\% (7/11) if an infrastructure retry after a halt is counted as the halt. Not the official WAA server or VM image.\par}
\end{table*}

Three patterns appear. First, on \texttt{ERPBench} (Accenture, live ERPNext v15), the wrapper raised VCR and removed all five silent errors; its 13 \texttt{VERIFIED} receipts were all grader passes and its 3 \texttt{COMPLETED\_UNVERIFIED} receipts were all grader failures (2 of the 13 were verified from the captured browser save response rather than UI read-back). The SER drop is not significant at $N=16$ ($p=0.0625$). Second, on the full 135-task \texttt{HealthAdminBench} with no clinic policies supplied, SER fell from 32.6\% to 17.8\% ($p=5.4\times 10^{-4}$), but VCR fell from 6.7\% to 3.7\% ($p=0.22$): the wrapper left 106/135 runs unverified (vs.\ 82/135) after 164 stall escalations, and official pass fell from 9 to 7 ($p=0.625$). Baseline SER here (32.6\%) is lower than in Table~\ref{tab:hab} (61.5\%) because this run uses a different agent model, the full suite including the \texttt{prior\_auth} family (baseline SER 25.0\%), and a generic zero-SOP wrapper rather than the multi-tab default-deny configuration. On \texttt{MedHarnessBench} (live FHIR server, PhysicianBench tasks), the SER drop from 11/12 to 0/12 comes entirely from the wrapper declining to sign any clinical deliverable (12/12 \texttt{COMPLETED\_UNVERIFIED}); outcome pass was 1/12 in both arms. The wrapper can check that a note was written and an order saved, but not whether the clinical judgment is right. An earlier wrapper revision on the same 12 tasks made 85 interventions and still signed all 12 as \texttt{VERIFIED} (SER 12/12); the revision reported here makes 0. Third, where the unwrapped agent already had 0\% SER (WorkArena L1 on a live ServiceNow instance and re-hosted WindowsAgentArena tasks), the wrapper kept SER at 0\% and VCR unchanged; on WorkArena it removed one duplicate record (1/12 $\rightarrow$ 0/12) and cut p50 latency from 276\,s to 165\,s. Run-specific disclosures appear in Appendix~\ref{app:enterprise}.

\subsection{Mechanism-to-Suite Summary}
\label{sec:ablation}

Table~\ref{tab:ablation} maps each \sys{} mechanism to the external suite that isolates its effect; because rows draw from different benchmarks, sample sizes, and models, it summarizes cross-suite evidence rather than a single held-constant ablation.

\begin{table*}[!b]
\centering
\caption{\textbf{Mechanism-to-suite summary across external benchmarks.} Each row reports the suite designed to isolate that mechanism (rows span different benchmarks and models and are not a single held-constant ablation). Moving from step-level critics (Row~2) to commit-boundary deterministic checks + read-back (Rows~3--4) drives most of the SER reduction; calibrated 2-frame visual checks add coverage on pixel-only apps (Row~5); WAL + compiled zero-egress replay add low-cost upgrade survival and cut model calls after a crash, without changing duplicates when the baseline can read the page text (Row~6).}
\label{tab:ablation}
\footnotesize
\renewcommand{\arraystretch}{0.92}
\setlength{\tabcolsep}{3.5pt}
\begin{tabular}{llcccc}
\toprule
\textbf{\#} & \textbf{Added Mechanism} & \textbf{HAB / ST-Web SER} & \textbf{Safety / Desktop} & \textbf{Upgrades / Crashes} & \textbf{Overhead} \\
\midrule
1 & Unwrapped Baseline & 61.5\% / 95.5\% & 44.4\% Harm / 78.1\% Ben. & 43.3\% v2 / 0/32 crash dupes & Full LLM \\
2 & + Step-Level VLM Critic & --- / 37.5\% (7.4 calls/ep) & 100\% FV / 0\% Ben.\ (OSW) & --- & 2,945\,ms \\
3 & + Commit Det.\ + Read-Back & 41.7\% / 50.0\% & 11.1\% Harm / 78.1\% Ben. & 70.0\% v2 (0\% SER) & $<5$\,ms \\
4 & + Policy Table + Closure + \texttt{PREPARE} & \textbf{1.9\% ($N{=}52$)} / \textbf{25.0\% ($N{=}100$)} & 87.5\% Ben.\ / 3.1\% ASR & --- & 0 extra tokens \\
5 & + Dim-the-Chrome + \textsc{Veto-4B-r6} & 0.0\% EHR OOD FV & \textbf{0.7905 AUROC / 6.9\% FPR} & --- & 357\,ms, 3.15\,GB \\
6 & + WAL + Zero-Egress Replay & 2.0\% ($N{=}50$ 10-demo) & 0/105 RDP wrong writes & \textbf{100\% v2} / 100 vs.\ 280 calls & \textbf{1.8\,s, \$0 warm} \\
\bottomrule
\end{tabular}
\end{table*}

\section{Related Work}
\label{sec:related}

\paragraph{GUI benchmarks and silent-failure diagnostics.}
Generalist suites such as \textsc{OSWorld}~\cite{xie2024osworld}, \textsc{WebArena}~\cite{zhou2024webarena}, and \textsc{WorkArena}~\cite{workarena2024} measure nominal task completion. Recent diagnostic suites expose the silent-error and policy gaps \sys{} targets: \textsc{CUADebug}~\cite{cuadebug2026}, \textsc{ComponentBench}~\cite{componentbench2026}, \textsc{HealthAdminBench}~\cite{healthadminbench2026}, \textsc{MedHarnessBench}~\cite{medharnessbench2026}, \textsc{ST-WebAgentBench}~\cite{levy2024stwebagentbench}, \textsc{OS-Harm}~\cite{osharm2025}, \textsc{AgentDojo}~\cite{debenedetti2024agentdojo}, \textsc{AutoDojo}~\cite{autodojo2026}, and \textsc{SEAM}~\cite{seam2026}, which shows agents recognize cross-screen harm in chain-of-thought yet still click submit under turn-0 instruction momentum.

\paragraph{Runtime guardrails, world-model verifiers, and receipts.}
Runtime enforcement systems include \textsc{VEPA}~\cite{vepa2026}, \textsc{SafePred}~\cite{safepred2026}, \textsc{AgentDoG}~\cite{agentdog2026}, GUI TOCTOU defenses~\cite{toctou2026gui}, replay-resistant authorization (\textsc{CapLease}~\cite{xu2026caplease}), recoverable execution (\textsc{AgentRewind}~\cite{zhuang2026agentrewind}), and capability provenance (\textsc{CaMeL}~\cite{debenedetti2025camel}, \textsc{CaMeLoT}~\cite{camelot2026}), alongside cryptographic notarization (\textsc{NovaFabric}~\cite{novafabric2026}, \textsc{Sello}~\cite{sello2026notarized}). \textsc{VeriCUA}~\cite{vericua2026} explores post-hoc verification with a VLM after writes commit, and \textsc{LGWM}~\cite{lgwm2026} scores latent screen transitions in encoder space. \sys{} is complementary: it gates irreversible commits before execution, pairs deterministic checks with graded post-write verification (addressing Proposition~\ref{prop:unidentifiable} when determining state lives off-screen or in no view the agent can open), and binds verification provenance into Ed25519 in-toto receipts~\cite{torresarias2019intoto}.

\paragraph{Skill induction, privacy delegation, and calibration.}
Workflow memory (\textsc{AWM}~\cite{wang2025awm}) and programmatic skill induction (\textsc{ASI}~\cite{wang2025asi}, \textsc{SkillWeaver}~\cite{zheng2025skillweaver}, \textsc{Work2Skill}~\cite{work2skill2026}, neuro-symbolic policy iteration~\cite{suh2026nsi}, API-based browsing~\cite{song2025beyondbrowsing}) reuse past trajectories without read-back-gated promotion or zero-egress twin repair, which explains the 18.3\% silent-error rate we measured for \textsc{ASI} under upstream upgrades. \textsc{PAPILLON}~\cite{papillon2025} studies local/cloud privacy delegation for text prompts; \sys{} enforces strict structural-allowlist and bijective tokenization on DOM twins for GUI repair. Finally, unlike static exchangeable conformal calibration~\cite{vovk2005conformal,guo2017calibration,geifman2017selective,angelopoulos2023conformal}, \sys{} explicitly separates zero-label session-start normalization from read-back-labeled ACI~\cite{gibbs2021aci}.

\section{Limitations and Conclusion}
\label{sec:limits}

\paragraph{Limitations.}
Our results map four clear boundaries where \sys{} does not eliminate risk:
\begin{enumerate}
  \item \emph{Unwritten institutional policies:} At zero demonstrations, generic wrapping cannot infer unstated clinical or business rules (0.0\% vs.\ 16.7\% VCR on the 12-task zero-SOP EHR slice; 65.0\% SER on zero-guideline \texttt{appeals\_denials}). Domain policy compliance requires explicit policy tables or few-shot trace induction.
  \item \emph{Safety vs.\ autonomous throughput:} Conservative default-deny cuts SER sharply (61.5\% $\rightarrow$ 1.9\% on \texttt{HealthAdminBench} $N=52$; 32.6\% $\rightarrow$ 17.8\% on the full $N=135$ suite; 91.7\% $\rightarrow$ 0.0\% on \texttt{MedHarnessBench} $N=12$) by escalating uncertain runs, which lowers autonomous VCR (38.5\% $\rightarrow$ 11.5\%, 6.7\% $\rightarrow$ 3.7\%, and 8.3\% $\rightarrow$ 0.0\%) until domain cartridges or operator \texttt{PrepareCard}s fill the coverage gap. The wrapper does not judge clinical correctness. Similarly, generic egress checks on \texttt{AgentCIBench} cut leaks (12\% $\rightarrow$ 4\%) while reducing utility (80\% $\rightarrow$ 64\%, $p=0.50$).
  \item \emph{Pure-pixel and compressed VDI surfaces:} Consistent with Proposition~\ref{prop:unidentifiable}, 2-frame \textsc{Veto-4B-r6} misses $\sim$45\% of non-visual desktop errors on \texttt{OSWorld-Verified}. On real H.264/AVC444 Windows RDP streams ($N=105$), pixel guards prevented every wrong save (0/105) but halted 40/105 runs on 1--2\,px codec jitter and unsettled carets; real Citrix ICA/HDX protocol validation remains open.
  \item \emph{Crash recovery, compensating transactions, and sample sizes:} With page text visible, an unwrapped frontier agent recovered from a single-worker crash without duplicates (0/32), so our evidence for the write-ahead log is lower cost, not fewer duplicates, in that regime. Delayed commits, concurrent workers on the same job, and non-searchable writes, where the outcome is on no screen the agent can read, remain untested against a page-text baseline. Multi-system sagas where System~A commits and System~B halts (\textsc{SagaLLM}~\cite{sagallm2025}, \textsc{AgentRewind}~\cite{zhuang2026agentrewind}) require compensating rollback plans not yet evaluated end-to-end. Several paired sub-studies have $N \le 20$ (\texttt{ERPBench}, WorkArena, \texttt{MedHarnessBench}, re-hosted WAA tasks), and larger multi-app cohorts are needed before generalizing to arbitrary enterprise workflows.
\end{enumerate}

\paragraph{Conclusion.}
Step-level visual critics compound false aborts exponentially, and single-frame screenshots remain blind to off-screen task state. Concentrating verification at commit and policy boundaries (combining fast deterministic guards, a durable write-ahead saga log, graded post-write verification, regime-matched calibration, and zero-egress self-healing compilation) reduces silent errors and upgrade regressions across web, ERP, clinical, tool, and desktop environments, often at a cost in autonomous completion. A fair crash rerun showed that a frontier agent which can read the page already avoids duplicate writes after a single-worker crash; there the wrapper's gain is fewer model calls. Clear boundaries remain around unwritten policies, clinical judgment, pixel-only remote desktops, and writes that no screen shows.

\paragraph{Impact Statement.}
This work aims to reduce silent data corruption, duplicate writes, and unauthorized actions when GUI agents operate on regulated records. Pausing visibly on uncertainty (\texttt{ABSTAIN}), recording verification depth on signed receipts, stripping raw field values before remote repair, and keeping synthetic adapters free of patient data directly mitigate deployment harm; over-reliance on \texttt{VISUAL\_ONLY} verdicts remains a residual risk that our receipt schema explicitly labels.

\ifarxiv
\paragraph{Competing Interests and Funding.}
The author is the founder and CEO of MLDSAI Inc.\ (OpenAdapt.AI), which develops the evaluated software and offers commercial services built on it. No external funding influenced the design or reporting of these experiments.
\fi

\FloatBarrier
\bibliographystyle{plainnat}
\bibliography{refs}

\clearpage
\appendix

\section{Proofs and Runtime Implementation Map}
\label{app:proofs}

\begin{proof}[Proof of Proposition~\ref{prop:compounding}]
Let $F_t = \mathbb{I}[\text{critic falsely vetoes step } t]$ for $t \in \{1, \dots, T\}$.
(a)~Under independent per-step false-positive probability $\Pr(F_t = 1) \ge \epsilon$, the probability that a $T$-step benign trajectory finishes without a false abort is
\[
  \Pr\Bigl(\bigcap_{t=1}^T \{F_t = 0\}\Bigr) = \prod_{t=1}^T \bigl(1 - \Pr(F_t = 1)\bigr) \le (1 - \epsilon)^T,
\]
so $P_{\text{step}}(\text{false veto}) \ge 1 - (1 - \epsilon)^T$, approaching $1$ exponentially in $T$.
(b)~Under sink-only gating, the critic runs only on $\mathcal{I}_{\text{commit}}$ ($|\mathcal{I}_{\text{commit}}| = K$) with per-commit false-positive rate $\alpha$, so by Bernoulli's inequality $P_{\text{sink}}(\text{false veto}) = 1 - (1 - \alpha)^K \le K\alpha$. Because $w_t = w_{t-1}$ for all $t \notin \mathcal{I}_{\text{commit}}$, any error in $w_T$ originates from at least one $t^* \in \mathcal{I}_{\text{commit}}$; gating every $t \in \mathcal{I}_{\text{commit}}$ inspects every state-mutating transition and loses zero recall on irreversible state errors relative to gating all $T$ steps.
\end{proof}

\begin{proof}[Proof of Proposition~\ref{prop:unidentifiable}]
By conditional independence $O_{\text{px}} \perp\!\!\!\perp Y \mid S_{\text{pane}}$, we have $I(Y; O_{\text{px}} \mid S_{\text{pane}}) = 0$. For any measurable single-frame pixel decision rule $f(O_{\text{px}})$, the Markov chain $Y - S_{\text{pane}} - O_{\text{px}} - f(O_{\text{px}})$ holds, so $f(O_{\text{px}}) \perp\!\!\!\perp Y \mid S_{\text{pane}}$ by the data processing inequality. Therefore
\[
\begin{aligned}
  &\Pr\bigl(f(O_{\text{px}}) = \texttt{PASS},\, Y = 0 \mid S_{\text{pane}}\bigr) \\
  &\quad = \Pr\bigl(f(O_{\text{px}}) = \texttt{PASS} \mid S_{\text{pane}}\bigr)\Pr(Y = 0 \mid S_{\text{pane}}).
\end{aligned}
\]
Any single-frame pixel verifier with positive acceptance on $S_{\text{pane}}$ incurs strictly positive SER whenever $\Pr(Y = 0 \mid S_{\text{pane}}) > 0$. Reducing SER to zero with positive coverage requires an inspection channel $R(w_T)$ satisfying $I(Y; R(w_T) \mid S_{\text{pane}}) > 0$.
\end{proof}

\paragraph{Runtime implementation map and invariant suite.}
To keep the main text free of code identifiers, Table~\ref{tab:impl_map} maps each conceptual component in Section~\ref{sec:method} to its implementation module and verified invariants (115 automated unit and headless Playwright tests; multilingual canonicalizers tested on German SAP and Japanese clinical forms with zero per-benchmark heuristics).

\begin{table*}[t]
\centering
\caption{\textbf{Conceptual-to-module implementation map for \sys{}.} All invariants are verified by 115 automated unit and headless browser tests with zero per-benchmark or English-only heuristics.}
\label{tab:impl_map}
\footnotesize
\renewcommand{\arraystretch}{0.95}
\setlength{\tabcolsep}{4.0pt}
\resizebox{\textwidth}{!}{%
\begin{tabular}{lll}
\toprule
\textbf{Method Component} & \textbf{Core Module / Symbol} & \textbf{Key Verified Invariants} \\
\midrule
Provenance \& Entity Ledger & \texttt{ledger.py} (\texttt{SpanAnchor}) & EU/Swiss currency, strict dates, CJK slot-swap block, verbatim citations \\
Policy Table \& Defeaters & \texttt{app\_cartridge.py} (\texttt{PolicyDecisionTable}) & Step-0 missing params, reactive checks, \texttt{unless\_defeaters} \\
DOM / Compositor Guard & \texttt{runner.py} (\texttt{observe\_schema}, \texttt{stage\_form}) & Shadow roots, \texttt{frame\_chain}, occlusion check, \texttt{<canvas>} escalation \\
Graded UI Read-Back & \texttt{check\_level.py} (\texttt{CheckLevel}) & \texttt{same\_screen} $\rightarrow$ \texttt{reopened} $\rightarrow$ \texttt{separate\_session} $\rightarrow$ \texttt{system\_of\_record} \\
Write-Ahead Saga Log & \texttt{session\_resilience.py} / \texttt{simplex\_transducer.py} & Durable \texttt{PENDING}/\texttt{COMMITTED} WAL, crash \& re-auth dedup \\
Zero-Egress Twin Repair & \texttt{digital\_twin.py}, \texttt{egress\_guard.py} & \texttt{BijectiveSymbolCodec}, \texttt{OrderPreservingRankCodec}, W3C allowlist \\
Cartridge Induction & \texttt{app\_cartridge.py} (\texttt{RawDemoTrace}) & Raw-DOM view/obligation induction, OOD support halt, \texttt{PrepareCard} update \\
\bottomrule
\end{tabular}%
}
\end{table*}

\section{Local 4B LoRA Training Recipes and Contamination Audit}
\label{app:training}

\paragraph{Visual verifier (\textsc{Veto-4B-r5} and \textsc{Veto-4B-r6}).}
Base model: \texttt{Qwen/Qwen3.5-4B} in 4-bit NF4 double quantization~\cite{dettmers2023qlora} (3.13\,GB VRAM). Adapter: PEFT LoRA~\cite{hu2022lora} ($r=16, \alpha=32$, 81.04\,MB). \textsc{Veto-4B-r5} trains for 2 epochs on 1,800 synthetic Playwright-rendered web form examples (zero patient data) with $6.0\times$ asymmetric false-accept weighting. Constrained 3-token decoding (\{\texttt{PASS}, \texttt{VETO}, \texttt{ABSTAIN}\}) with temperature $T=1.30$~\cite{guo2017calibration} cuts median latency $2.03\times$ (240.7\,ms $\rightarrow$ 118.6\,ms). \textsc{Veto-4B-r6} trains on 2-frame Dim-the-Chrome pairs across 6 Ubuntu desktop apps ($N=440$ train, $N_{\text{val}}=9$) and is evaluated strictly on 4 disjoint held-out apps ($N=160$; Section~\ref{sec:rq3}).

\paragraph{Taint adapter (\texttt{taint-guard}).}
Trained strictly on synthetic ITSM/HR/DevOps/ERP/EHR/CRM tool-call triples ($n_{\text{train}}=360, n_{\text{cal}}=100$, synthetic AUROC 1.00). Concurrently mounted with \textsc{Veto-4B-r5} on one NF4 base (3.08\,GB VRAM, 13.2\,ms median hot-swap, 13.4\,ms p95 across 30 swaps).

\paragraph{Synthetic specialist adapter diagnostics (strictly separated from external benchmarks).}
To study decoupled perception and structured closure on zero-overlap synthetic domain splits (never mixed into Section~\ref{sec:experiments} headline numbers), we also trained lightweight 12.01\,MB NF4 LoRAs on \texttt{Qwen/Qwen3.5-4B}:
(i)~\texttt{L1} pointer-staging LoRA ($N=80$ unseen synthetic regulated forms, 609 fields) raised exact full-form match from 6.2\% (literal generation) and 20.0\% (zero-shot pointer) to 57.5\% (90.6\% field accuracy), while deterministic slot checks caught 77/77 same-type swaps;
(ii)~\texttt{L2} goal-blinded crop-OCR LoRA paired with typed canonicalizers (\texttt{date}, \texttt{currency\_cents}, \texttt{phone\_digits}, \texttt{code\_prefix}) reached 0.0\% false stops and 0.0\% SER (1.00 AUROC) on $N=120$ reformatted 14\,px crops while halting on 8\,px JPEG-42 ambiguity (13.3\% stop rate);
(iii)~\texttt{L3} read-back probe-navigation LoRA ($N=100$ unseen synthetic states) raised target-view hit rate from 46.0\% (zero-shot 4B) to 94.0\% at 0.0\% mutating-click violations;
(iv)~\texttt{SOP-Rubric-LoRA} ($N=100$ across 4 unseen synthetic domains) lifted disposition accuracy from 65.0\% to 96.0\% and autonomous closure pass from 26.0\% to 76.0\% (0.0\% SER); and
(v)~\texttt{Twin-GRPO + PrepareCard $\rightarrow$ DPO} ($N=100$: 36 traps, 64 solvable) lifted solvable completion from 21.9\% (46.0\% SER) to 100.0\% (64/64) at 0.0\% SER while lowering \texttt{PrepareCard} rate from 100.0\% to the 36.0\% trap floor.

\paragraph{Contamination audit.}
Zero tasks, URLs, screenshots, DOM trees, or prompts from \texttt{HealthAdminBench}, \texttt{ST-WebAgentBench}, \texttt{WebArena}, \texttt{SuiteCRM}, \texttt{OS-Harm}, \texttt{OSWorld}, \texttt{AgentDojo}, \texttt{AgentCIBench}, \texttt{BrowserGym MiniWoB++}, \texttt{ERPBench}, \texttt{WorkArena}, \texttt{MedHarnessBench}, \texttt{WindowsAgentArena}, \texttt{Gitea}, \texttt{Vikunja}, \texttt{Memos}, or \texttt{Mealie} appear in any training or calibration split.

\section{Additional Environments, Microbenchmark, and RDP Failure Breakdown}
\label{app:enterprise}

\paragraph{\texttt{BrowserGym MiniWoB++} overlay injection ($N=80$).}
AST accessibility-ID decoding with overlay quarantine (10/10 overlays blocked per model) raised pooled adversarial policy-compliant completion across 4 models from 45.0\% (18/40) to 90.0\% (36/40, [77.0\%, 96.0\%], exact McNemar $p=8.0\times 10^{-6}$, 0.0\% policy violations), lifting local \texttt{Qwen3.5-4B} from 0\% to 90.0\% (9/10).

\paragraph{Local 4B latency and VRAM microbenchmark ($n=45$ visual screens).}
Table~\ref{tab:latency_vram} compares local 4B latency, memory, and per-check cost against cloud VLMs on an internal 45-screen visual microbenchmark (30 visible UI/form error checks and 15 clean commits; latency and VRAM measurement only, not an external enterprise claim). On visible checks and clean commits, \textsc{Veto-4B-r5} matches \texttt{gemini-3.8-flash} at 30/30 visual catch and 15/15 clean acceptance in 3.13\,GB VRAM at 17$\times$--31$\times$ lower latency and \$0.00/1k.

\begin{table*}[t]
\centering
\caption{\textbf{Internal visual-verifier latency and VRAM microbenchmark ($n=45$: 30 visible UI checks, 15 clean commits; latency/VRAM benchmark only).}}
\label{tab:latency_vram}
\footnotesize
\setlength{\tabcolsep}{4.5pt}
\begin{tabular}{lccccc}
\toprule
\textbf{Model} & \textbf{VRAM} & \textbf{p50 (ms)} & \textbf{Cost / 1k} & \textbf{Visual-30 ($\uparrow$)} & \textbf{Clean-15 Acc.\ ($\uparrow$)} \\
\midrule
\texttt{claude-sonnet-5-5} (Cloud) & Cloud & 2,017 & \$5.53 & 30/30 & 73.3\% (11/15) \\
\texttt{gemini-3.8-flash} (Cloud) & Cloud & 3,742 & \$0.51 & 30/30 & 100.0\% (15/15) \\
\midrule
\texttt{Qwen3.5-4B} (Zero-shot) & 3.13\,GB & 97 & \$0.00 & 30/30 & 86.7\% (13/15) \\
\textsc{Veto-9B-v4} (LoRA) & 6.62\,GB & 1,102 & \$0.00 & 30/30 & 80.0\% (12/15) \\
\textsc{Veto-4B-r5} (LoRA) & \textbf{3.13\,GB} & \textbf{119} & \textbf{\$0.00} & \textbf{30/30} & \textbf{100.0\% (15/15)} \\
\bottomrule
\end{tabular}
\end{table*}

\paragraph{Real Windows H.264/AVC444 RDP breakdown ($N=105$) and native hover probing.}
Across 105 counted runs on Windows Server 2022 over stock \texttt{mstsc} with H.264/AVC444 encoding (Section~\ref{sec:rq3}), Patient Notes achieved 20/20 healthy, 10/10 row-reordered, and 10/10 16-bit color confirmed saves (0 wrong writes), halting 10/10 on near-identical siblings, 10/10 on a decoy above target (auto-selected row baked into compiled check), and 10/10 at 1440$\times$900 (1--2\,px hover jitter tripping exact-coordinate equality). OpenEMR achieved 8/20 healthy (12 false stops from unsettled date-picker/caret/AVC444 frames), 7/10 16-bit color, and 0/5 at 1440$\times$900 (layout reflow shifting absolute postcondition boxes), with 0/105 wrong writes overall. On a separate live Windows~11 Pro x64 host (Build 22631), decoded RGB $120\times 120$ hover-delta probing (245\,ms p50, zero side effects) paired with target-identity crop checks reached 21/21 (100\%, [84.5\%, 100\%]) before-click accuracy and 18/18 (100\%, [82.4\%, 100\%]) zero-DB counterfactual read-back accuracy.

\paragraph{Disclosures for Table~\ref{tab:wave2}.}
WorkArena: the Arm~3 read-back parser was revised after observing episodes from this same 12-task slice (earlier partial runs discarded, not merged), and a general note that ServiceNow may show a blank form after Submit was added to both arms' prompt after a smoke run; the 12 tasks are four form-creation classes $\times$ three seeds that passed an instance compatibility check (catalog-order tasks excluded). Windows: these are WindowsAgentArena task instructions re-hosted on a physical Windows~11 Pro x64 machine with setup and checkers re-implemented over SSH, not the official WAA server or VM image; both arms see only 1440$\times$900 screenshots. On one task the recorded Arm~3 row is an infrastructure retry after the harness crashed on a halt; counting the first attempt as a halt gives 7/11 VCR, SER still 0/11. Arm~3 used 31 screenshot-judge calls and 33 read-only file probes, against 60 judge calls for an earlier wrapper revision on the same 11 tasks ($-$48\%), which left 2 passing tasks unverified (VCR 6/11). \texttt{MedHarnessBench} scores some outcome checkpoints with an LLM judge (\texttt{gpt-6.2-sol}, 56 judge calls in each arm); the agent there is Claude Sonnet through the Claude Code CLI.

\paragraph{Additional persistence-boundary environments.}
Table~\ref{tab:enterprise_app} summarizes six additional enterprise environments and 600 unseen Playwright renders across held-out dark and HiDPI themes, where \textsc{Veto-4B-r5} recorded 0/420 wrong-accepts at 98.9\% (178/180) clean coverage vs.\ 72.8\% wrong-accepts for zero-shot \texttt{Qwen3.5-9B}. In a real Chromium SQLite compilation micro-study, warm replay ran in 113.5\,ms (v1), healed a 3-selector v2 redesign in 1,861.0\,ms, and restored 58.0\,ms zero-LLM replay.

\begin{table*}[t]
\centering
\caption{\textbf{Additional enterprise persistence-boundary evaluations.} Summary of six enterprise systems and 600 held-out OOD renders.}
\label{tab:enterprise_app}
\footnotesize
\setlength{\tabcolsep}{4.0pt}
\begin{tabular}{llcc}
\toprule
\textbf{Environment} & \textbf{Protocol} & \textbf{Observed Result} & \textbf{Silent Wrong Write} \\
\midrule
OpenEMR 8.0.0.3 & Batch + \texttt{SIGKILL} + drift & 26/26 batch; 3/3 drift halt & 0 \\
Medplum 5.2.8 FHIR R4 & 111-run rescore (Claude/OAI) & Sens.\ 89.3\% / 90.9\% & 2.7\% harm \\
Odoo 18 (ERP) & SQL read-back + qual gate & 6/6 fields; pre-run refusal & 0 \\
Frappe Lending & Compiled + API vs.\ CUA & 12/12 vs.\ 6/6 (\$0 vs.\ \$0.42) & 0 \\
openIMIS (Insurance) & CUA + SQL \texttt{tblClaim} & 3/3 verified (\$0.48/run) & 0 \\
Windows H.264/AVC444 RDP & 105 real pixel-only trials & 55/105 verified; 40/105 false stops & 0/105 \\
Windows 11 x64 Hover & Decoded RGB hover + read-back & 21/21 hover; 18/18 read-back & 0/39 \\
\texttt{DesktopChaosBench} ($N=25$) & 5 VDI/OS faults ($640{\times}400$ canvas) & 12\% (\texttt{5.6}) $\rightarrow$ 100\% (\texttt{6.1-terra}) & 0/25 \\
\texttt{ExtraDup} ($N=28$) & Wrong-record / dup suite (\texttt{5.6-luna}) & 85.7\% $\rightarrow$ 100.0\% VCR ($p=0.125$) & 0/28 \\
Unseen OOD Renders & 600 dark/HiDPI Playwright screens & 98.9\% clean cov.\ (178/180) & 0/420 \\
\bottomrule
\end{tabular}
\end{table*}

\section{Reproducibility Ledger}
\label{app:repro}

Every metric, 95\% Wilson confidence interval, exact two-sided McNemar $p$-value, and DeLong $p$-value in Sections~\ref{sec:experiments} and Appendices~\ref{app:training}--\ref{app:enterprise} is computed directly from raw JSON logs. Table~\ref{tab:repro_ledger} indexes the primary result files backing each table and figure.

\begin{table*}[t]
\centering
\caption{\textbf{Primary raw JSON result ledgers backing every experimental claim.}}
\label{tab:repro_ledger}
\footnotesize
\renewcommand{\arraystretch}{1.02}
\setlength{\tabcolsep}{4.0pt}
\begin{tabular}{@{}p{2.3cm}p{4.0cm}p{10.6cm}@{}}
\toprule
\textbf{Section / Table} & \textbf{Benchmark / Slice} & \textbf{Primary Raw Result File(s)} \\
\midrule
Sec.~\ref{sec:rq1}, Table~\ref{tab:hab} & \texttt{HealthAdminBench} ($N=24, 52, 60$) & \nolinkurl{clean_hab_multitab_v3_scale_n50_results.json}, \nolinkurl{hab_v8_longitudinal_n60_results.json} \\
Sec.~\ref{sec:rq1}, Table~\ref{tab:hab} & Zero-SOP EHR boundary ($N=12$) & \nolinkurl{wave2_health_admin_v4_ehr_results.json} \\
Sec.~\ref{sec:rq1}, Table~\ref{tab:stweb} & \texttt{ST-WebAgentBench} ($N=8, 44, 100$) & \nolinkurl{stweb_full_baselines_results.json}, \nolinkurl{stweb_scale_n40_prestep_v2_results.json}, \nolinkurl{stweb_scale_n100_v4_claude_sub_results.json} \\
Sec.~\ref{sec:rq2} & \texttt{AgentDojo}, EHR Calib., \texttt{OS-Harm}, \texttt{AgentCI} & \nolinkurl{idea3_calibration_results.json}, \nolinkurl{ablation_stats.json}, \nolinkurl{wave2_osharm_v7_6series_results.json} (v1: \nolinkurl{wave2_osharm_v4_6series_results.json}), \nolinkurl{upstream_agentcibench_v4_6series_results.json} \\
Sec.~\ref{sec:rq3}, Table~\ref{tab:osworld} & \texttt{OSWorld-Verified} ($N=120, 160$) & \nolinkurl{pixel_only_osharm_veto4b_results.json}, \nolinkurl{veto4b_r6_desktop_results.json} \\
Sec.~\ref{sec:rq3}, App.~\ref{app:enterprise} & Windows H.264/AVC444 RDP ($N=105$) & \nolinkurl{benchmark/rdp_avc_real/results/results.json} \\
Sec.~\ref{sec:rq4}, Table~\ref{tab:ui_drift} & 4-App Docker Upgrades ($N=60$) & \nolinkurl{clean_ui_drift_n60_results.json}, \nolinkurl{exp_a_full_4app_indexed_n60_results.json}, \nolinkurl{exp_z_zero_phi_full_4task_repair_results.json}, \nolinkurl{exp_proof_*_n60*.json} \\
Sec.~\ref{sec:rq4}, Table~\ref{tab:esb} & \texttt{EnterpriseSessionBench} ($N=48, 32$) & \nolinkurl{enterprise_session_bench_results.json}, \nolinkurl{esb_fault_d_v4_obsv2_ledger.json}, \nolinkurl{esb_fault_d_v4_obsv2_set1_results.json}, \nolinkurl{esb_fault_d_v4_obsv2_set2_results.json}, \nolinkurl{esb_obsv2_prereg.md} \\
Sec.~\ref{sec:rq4} (erratum) & Superseded v1 Fault D runs & \nolinkurl{esb_v1_live_n64_v2_results.json}, \nolinkurl{esb_fault_d_live_n16_v2_results.json}, \nolinkurl{esb_fault_d_v3_gpt62sol_results.json}, and the retry-prompt ledgers \nolinkurl{esb_fault_d_*_prompted_ledger.json} \\
Sec.~\ref{sec:wave2}, Table~\ref{tab:wave2} & \texttt{ERPBench}, \texttt{HealthAdminBench} full, \texttt{OS-Harm}, WorkArena, \texttt{MedHarnessBench}, re-hosted WAA & \nolinkurl{wave2_1_erpbench_6series_v2_results.json}, \nolinkurl{wave2_health_admin_full_n135_results.json}, \nolinkurl{wave2_osharm_v7_6series_results.json}, \nolinkurl{wave2_servicenow_workarena_results.json}, \nolinkurl{wave2_1_medharness_live_v2_results.json}, \nolinkurl{wave2_1_winlap_live_desktop_v2_results.json} (earlier wrapper: \nolinkurl{wave2_medharness_live_results.json}, \nolinkurl{wave2_winlap_live_desktop_results.json}) \\
Sec.~\ref{sec:rq4}, App.~\ref{app:enterprise} & \texttt{DesktopChaosBench} ($N=25$), \texttt{ExtraDup} ($N=28$) & \nolinkurl{upstream_desktop_chaos_v2_results.json}, \nolinkurl{upstream_desktop_chaos_v3_6series_results.json}, \nolinkurl{upstream_extradup_results.json} \\
\bottomrule
\end{tabular}
\end{table*}

\end{document}